\documentclass{article}

\usepackage{arxiv}

\usepackage[utf8]{inputenc} 
\usepackage[T1]{fontenc}    
\usepackage[pagebackref,breaklinks,colorlinks]{hyperref}
\usepackage{url}            
\usepackage{booktabs}       
\usepackage{amsfonts}       
\usepackage{nicefrac}       
\usepackage{microtype}      
\usepackage{lipsum}

\usepackage{amssymb}
\usepackage{amsmath}
\usepackage[normalem]{ulem}
\usepackage{multirow}
\usepackage{array}
\usepackage{graphicx}

\title{Latent Graph Attention for Enhanced Spatial Context}

\author{
    Ayush Singh \\
    Department of Mathematics and Computing \\
    Indian Institute of Technology, ISM Dhanbad \\
    Dhanbad, India \\
    \texttt{ayush.singh.222k@gmail.com} \\
    \And
    Yash Bhambhu \\
    Department of Mathematics and Computing \\
    Indian Institute of Technology, ISM Dhanbad \\
    Dhanbad, India \\
    \texttt{yashbhambhu.18je0949@am.iitism.ac.in}
    \And
    Himanshu Buckchash \\
    Department of Computer Science \\
    UiT The Arctic University of Norway \\
    Tromsø, Norway \\
    \texttt{himanshu.buckchash@uit.no} \\
    \And
    Deepak K. Gupta \\
    Department of Computer Science \\
    UiT The Arctic University of Norway \\
    Tromsø, Norway \\
    \texttt{deepak.k.gupta@uit.no}
    \And
    Dilip K. Prasad \\
    Department of Computer Science \\
    UiT The Arctic University of Norway \\
    Tromsø, Norway \\
    \texttt{dilip.prasad@uit.no} \\
}

\begin{document}
\maketitle

\begin{abstract}
Global contexts in images are quite valuable in image-to-image translation problems. Conventional attention-based and graph-based models capture the global context to a large extent, however, these are computationally expensive. Moreover, the existing approaches are limited to only learning the pairwise semantic relation between any two points on the image. In this paper, we present Latent Graph Attention (LGA) a computationally inexpensive (linear to the number of nodes) and stable, modular framework for incorporating the global context in the existing architectures, especially empowering small-scale architectures to give performance closer to large size architectures, thus making the light-weight architectures more useful for edge devices with lower compute power and lower energy needs. 
LGA propagates information spatially using a network of locally connected graphs, thereby facilitating to construct a semantically coherent relation between any two spatially distant points that also takes into account the influence of the intermediate pixels.
Moreover, the depth of the graph network can be used to adapt the extent of contextual spread to the target dataset, thereby being able to explicitly control the added computational cost. To enhance the learning mechanism of LGA, we also introduce a novel contrastive loss term that helps our LGA module to couple well with the original architecture at the expense of minimal additional computational load.
We show that incorporating LGA improves the performance on three challenging applications, namely transparent object segmentation, image restoration for dehazing and optical flow estimation.
\end{abstract}


\section{Introduction}
\label{sec:intro}
Recent advancements related to convolutional neural networks (CNNs) have led to significant advancements in image-to-image translation problems. Some popular examples include edge-region based segmentation \cite{276131}, image restoration with neural network initialized using hand-crafted features \cite{UlyanovVL17}, dehazing with dark channel prior \cite{5206515}, among others. This success can be attributed to the capability of CNNs to extract deep and complex features from the images without needing any hand-crafted features.

\begin{figure}[t]
\centering
\includegraphics[width=\linewidth]{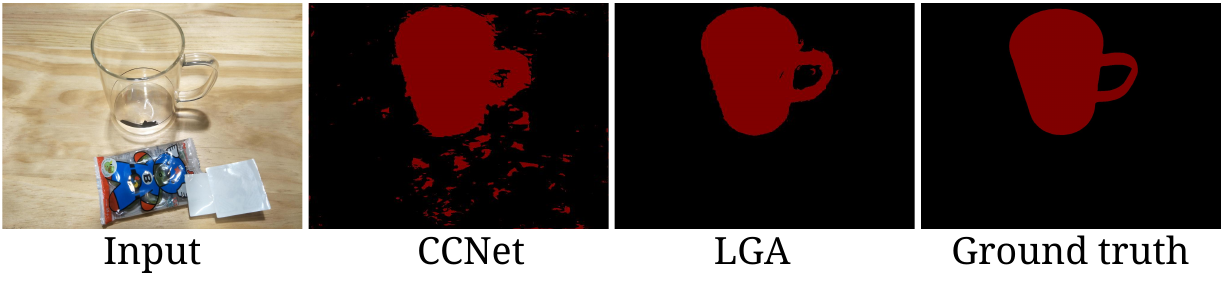}
\caption{Segmentation results obtained for a transparent object using CCNet \protect\cite{huang2020ccnet} and our method (LGA). Note that due to the lack of semantic coherence between far away points, CCNet produces partially incorrect segmentation.}
\label{fig:patch_propagation}
\end{figure}

The underlying mechanism of CNNs is that they reduce the spatial information contained in the images through gradually down-sampling to a set of feature maps. For this purpose, CNNs employ convolutional kernels that convolve pixels only locally without taking the whole global context into consideration. This works well for conventional downstream tasks such as classification, lack of global context limits their performance on challenging image-to-image translation problems such as segmentation of transparent objects \cite{mei2022large} and dehazing \cite{zhang2021hierarchical}.
In transparent objects, although the edges may bear a clear signature of the object, there are strong features of other background objects which adversely affect the segmentation process. To tackle this issue, it is desired that the information from the edges can be transferred to the other parts of the object as well.
In dehazing, shapes and edges of the object need to be restored from original images where hazing fuzzifies these features, and information on the non-local context should help. 

Several previous works exist on incorporating the global spatial context into the learning process. Some examples include layer-wise average feature augmentation \cite{parsenet}, using mixtures of conditional random fields and CNNs \cite{FCCRF,ETECRF} and constructing encoder-decoder style fully-convolutional networks which use deconvolutions to create output building upon U-Net \cite{unet}. Two popular approaches to introduce the global context without compromising the localization accuracy are the use of atrous or dilated convolutions \cite{deeplabv1, DeepLabv3} and transformer networks \cite{DBLP:journals/corr/abs-1804-05091}. Although both approaches have been successful, these are accompanied by significant increase in the number of parameters and the associated FLOPs. Alternatively, crisscross attention (referred as CCNet) \cite{huang2020ccnet} presents a light-weight solution to model the global context in the learning process. It collects contextual information from vertical and horizontal crisscross channels and applies an affinity operation in the latent space. Compared to the other approaches, this method is computationally efficient with a time complexity of $\mathcal{O}(N^{1.5})$ where $N$ denotes feature map size. 

Figure \ref{fig:patch_propagation} presents an example segmentation result obtained using CCNet. Interestingly, the capability of capturing the global context plays an adverse role in this case. CCNet relies on the semantic relation between any two distant points, however, it does not take into account the characteristics of the intermediate points between them. Clearly, as seen in Fig. \ref{fig:patch_propagation} parts of other objects getting segmented falsely as our object of interest. To circumvent this issue, we propose an alternate scheme that constructs \textit{chained attention} spatially, thereby taking into account the information of the entire route to compute the relation between any two distant pixels. This is achieved through the use of locally connected graph networks constructed over the latent feature maps, and these gradually propagate information to the distant neighbours of any point through a stacked set of layers used in the graph network.
Our approach is adaptive and provides the flexibility of choosing the desired spatial extent to be captured. Further, our approach is computationally faster and provides a speedup of $\sqrt{N}$ over CCNet.

\paragraph{Contributions}The main contributions of this paper can be summarized as follows:
\begin{itemize}
\itemsep0em
\item We present Latent Graph Attention (LGA), a graph-network based module to incorporate the global context in the existing CNN architectures.
\item LGA is computationally inexpensive, and it provides a speedup of $\sqrt{N}$ over the previous methods. The time complexity of LGA scales linearly with the number of connected nodes in the graph.
\item For stable and responsive learning, we introduce an LGA-specific contrastive loss term that amplifies the discrimination between the foreground and background and accordingly helps to learn the suitable weights for each edge of the graph.
\item We experimentally demonstrate that our LGA module when plugged into a small-scale architecture, helps to boost their performance with only minimal additional computational load. This empowers the development of efficient and compact architectures for edge devices.
\item We experimentally demonstrate the efficacy of LGA over a variety of hard image-to-image translation tasks, which include segmentation of transparent objects, image dehazing, and optical flow estimation.
\end{itemize}

\section{Related Work}
\paragraph{Context aggregation} Spatial context provides valuable information to many image-to-image translation tasks \cite{chen2017deeplab,peng2017large,wang2018non}. Context aggregation is a recurrent theme that appears in different types of convolution strategies, pooling variations, attention mechanisms, channel operations, architectural search \cite{DeepLabv3,zhao2017pyramid,huang2020ccnet,zhang2022resnest,liu2019auto}.
Dilated convolution, multi-scale feature generation, and large kernel approaches leverage contextual information based on modulation of receptive field to extract information from non-local regions \cite{zhao2017pyramid,ding2022scaling,chen2018searching,DeepLabv3,yang2018denseaspp}. Architecture modifications like Inception models, ResNeSt, SK-Net focus on multi-branch representation for splitting channel-wise information \cite{zhang2022resnest,li2019selective,szegedy2015going,szegedy2016rethinking}. Architecture search approaches Auto-DeepLab, Global2Local, DCNAS 
work on automating context modeling by searching multiple paths and channels at cell or network levels \cite{liu2019auto,gao2021global2local,zhang2021dcnas}.
Attention has served as a very versatile context aggregation mechanism, working at various scales like local \cite{harley2017segmentation}, non-local \cite{wang2018non}, global \cite{peng2017large}, cross or self-attention \cite{huang2020ccnet,petit2021u}. Recent trends have shifted from variants of attention to full attention in transformers \cite{dosovitskiy2020image,translab2} and its more generalized forms \cite{lu2021container,wang2021pyramid}.
However, this push towards context aggregation has come at the cost of network size and resources \cite{huang2020ccnet,Squeeze1,zhang2017shufflenet}. These challenges demand confluence of high-performing attention traits into small-size networks.

\paragraph{Light-weight architectures} Architectural buoyancy is generally achieved through small scale architectures \cite{Squeeze1,sandler2018mobilenetv2,zhang2017shufflenet}, or lighter attention schemes \cite{huang2020ccnet,li2020lightweight}, or distillation strategies \cite{qin2022multi,kong2022mdflow}. Of these three, an attention approach like Criss-Cross \cite{huang2020ccnet} is easily adaptable, highly modular \& task agnostic \cite{huang2020ccnet}. Criss-Cross attention (or CCNet) is one of the best light attention modules, however, it employs spatial position agnostic correlation, which does not work well when a sequential context propagation is required. Graph-based methods provide control over these pathways for devising chained attention mechanism \cite{co-segmentation,NIPS2016_04df4d43,lu2019graph}.

\begin{figure}
	\centering
	\includegraphics[width=.63\linewidth]{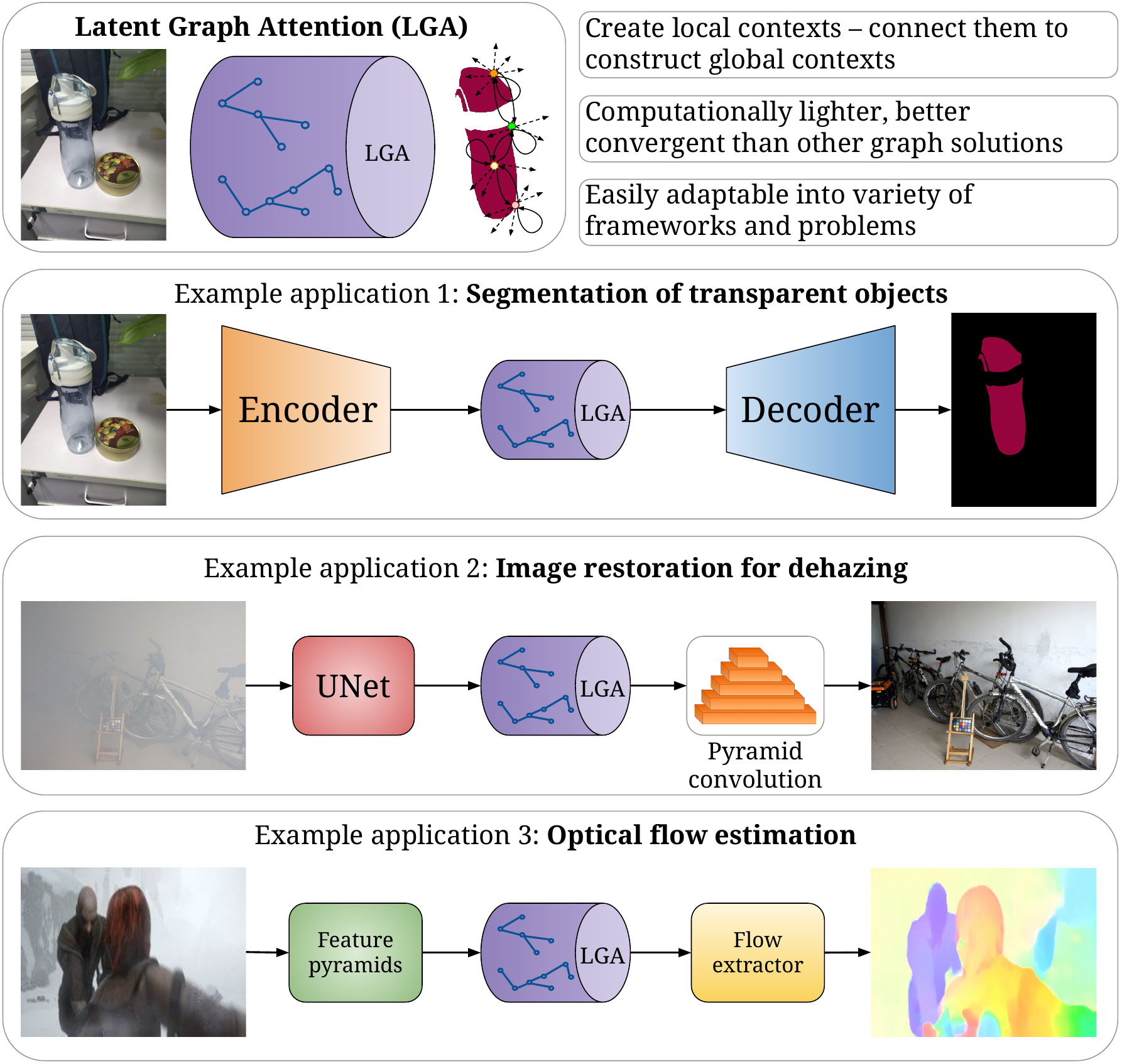}
	\caption{We present a novel concept of Latent Graph Attention (LGA), which can be easily integrated in wide variety of applications and architectures. Three challenging and open problems are considered as example applications of LGA in this article.}
\label{fig:teaser}
\end{figure}

\paragraph{Graph-based networks} Despite developments in graph-based solutions like generalization of CNNs as graphs \cite{NIPS2016_04df4d43,meng2021bi,zhang2019dual}, certain issues still remain. Banerjee \emph{et al.} ~\cite{co-segmentation}, proposed an end-to-end graph CNN method for transparent object segmentation by forming undirected edges between any two adjacent super-pixels. Although, unlike directed graphs, undirected graphs are unable to associate two far-away nodes to the same structural context. Moreover, the adjacency matrix generation in \cite{co-segmentation} for an image resulting in $N$ nodes bears a time complexity of $N^2$. This shows that achieving global context through graphs has high time complexity. To provide a more local context, Lu \emph{et al.} ~\cite{lu2019graph} incorporated graphs in a fully connected network (FCN) for better segmentation. They considered FCN features located $l$ distance away from each other as graph nodes. However, to have a larger context their approach incurs higher space-time complexity.

We present Latent Graph Attention (LGA) that overcomes the challenges outlined above. LGA employs outward directionality in the adjacency matrix through which it propagates information to its neighbouring nodes. When LGA layers are stacked together multiple times, the information reaches outward from each node to the distant nodes, thereby providing a non-local spatial context.

\section{The proposed approach}

\begin{figure}[!t]
\centering
\includegraphics[width=1\linewidth]{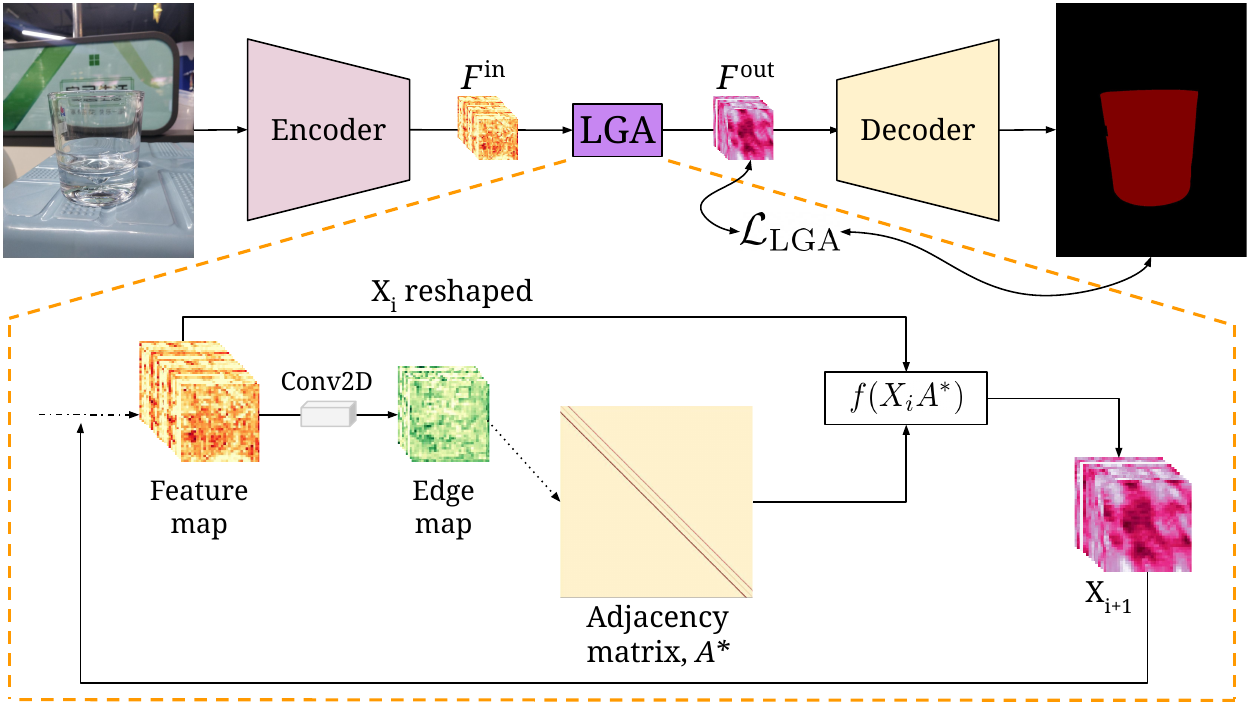}
\caption{Operations performed by the LGA layer are shown in the expanded view. Starting with the encoder's feature maps, $F^{\rm in}$, edge maps are created using 2D convolution. The dotted arrow between \emph{Edge maps} and \emph{Adjacency matrix} implies that this happens only once even if LGA layer is repeated multiple times. Next, the normalized adjacency matrix is used to calculate the output, $X_{i+1}$, for the $i^{th}$ LGA layer. The LGA contrastive loss, $\mathcal{L}_{\rm LGA}$, is computed between the output of LGA module, $F^{\rm out}$, and the ground truth. $X_{i+1}$ becomes the input to $i$+$1^{th}$ LGA layer.}
\label{fig:model}
\end{figure}

In this section, we present the concept of LGA, discuss its stability, and its modular adoption across architectures.

\subsection{Latent Graph Attention (LGA)} \label{sec:LGA}

LGA is a computationally inexpensive graph-based attention network designed to improve the contextual information for any given image. It constructs chained attention spatially through the use of multiple layers of the graph, thereby taking into account the information of the entire route when computing the relation between two distant pixels in the latent space. LGA directly operates on the latent feature map and produces additional attention maps which can directly be concatenated with the original feature map. Due to its simplistic nature, LGA can be incorporated with minimal effort in the existing architectures. Figure \ref{fig:teaser} shows how LGA can be plugged into some popular yet challenging image-to-image translation tasks.

\begin{figure*}[t]
	\centering
	\includegraphics[width=1\linewidth]{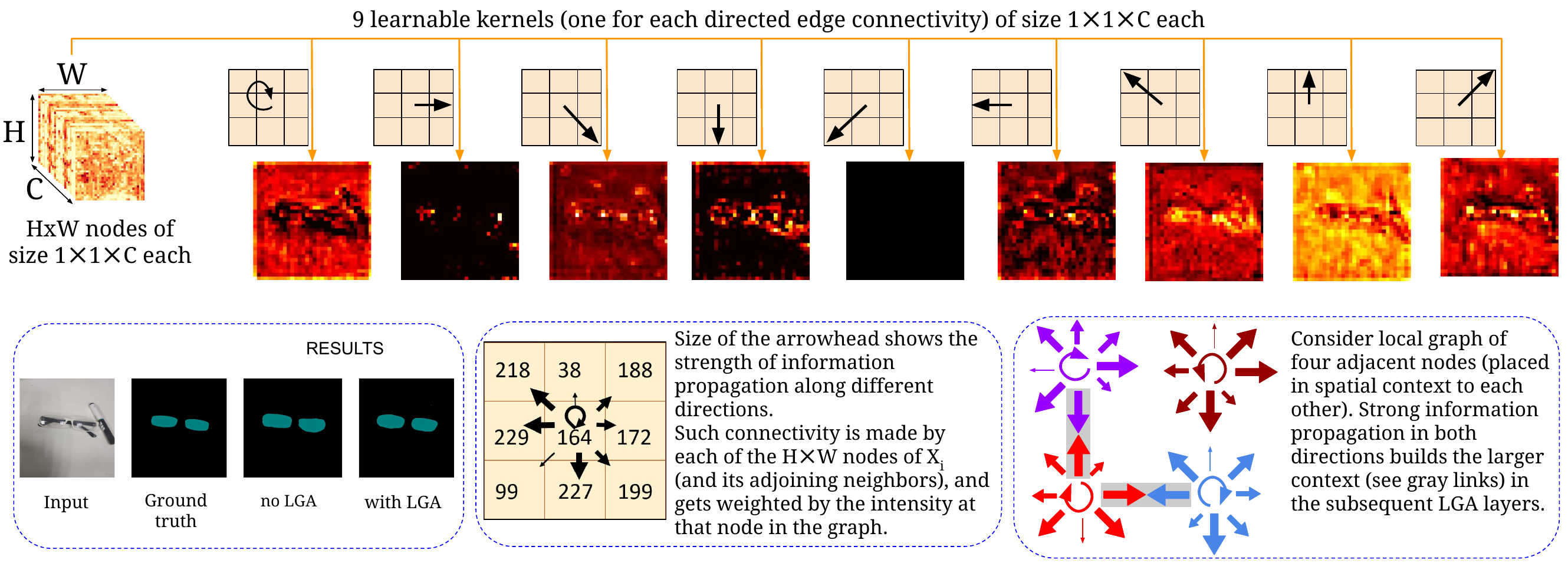}
	\caption{Top row shows the conversion of feature map $X_i$ to graph nodes. Each cell of $X_i$ in $H \times W$ plane (at all depths) is considered as one node in the graph. The graph information is extracted by 9 kernels one for each direction. Edge maps (in top row) are shown corresponding to the input, gt, predicted triplet (bottom-left). Mid-bottom figure shows how the node intensities represent the edge weights. Bottom-right figure shows how connectivity strength information propagates through recursive LGA layers.}
	\label{fig:feat_node}
\end{figure*}

\paragraph{Architecture of LGA} Fig. \ref{fig:model} presents a schematic overview of the working of our LGA module.  
It takes as input a feature map \mbox{$F^{\rm in} \in \mathbb{R}^{H \times W \times C}$}, where $H$, $W$ and $C$ denote the height, width and number of channels respectively. It constructs a graph using this feature map (as shown in Fig. \ref{fig:feat_node}). Every spatial feature on the feature map, with dimensions of $1 \times 1 \times C$, is considered a node of the graph thereby leading to $N=HW$ nodes in the entire graph, as shown in Fig. \ref{fig:feat_node}.

\textit{Learning the weights of the graph edges. }Depending on the directions in which strong spatial relation is to be learnt, edges should be constructed in the graph. We construct edges only between the immediate spatial neighbors based on a 8-connectivity pattern (see Fig. \ref{fig:feat_node}). This leads to 9 connections per node - one self-connection and 8 with the neighbors. The weights of the edges are learnable and each of these weights is generated using a 1-layered CNN comprising a $1 \times 1$ conv layer. Since we have 9 edges, there are a total of 9 such mini-networks that learn the edges.

\textit{Construction of adjacency matrix. }The learnt edges are used to construct an adjacency matrix $A$. It is a $N \times N$ matrix where $A_{ij} \in A$ denotes the weight of the edge connecting the $i^{\text{th}}$ node with the $j^{\text{th}}$ in the graph. For the sake of convergence, $A$ is normalized to form $A^*$. The normalization ensures that the feature vectors of all the nodes remain within a unit polysphere and are of comparable strengths with respect to each other. We define $A^* = AD^{-1}$, where \mbox{$D_{ij} = \sum_j {A_{ij}} + \epsilon$}. Here, $\epsilon$ is added to avoid the zero division. The matrix $A$ is normalized based on the weights of the outgoing edges since it gives better results and stability over the training process. 

\textit{Message passing. }With this construction, LGA facilitates message passing between the immediate neighbors in the latent space. To propagate information further away, multiple such graphs are stacked to build a graph network that constitutes the LGA module. The information propagation at each LGA layer (Fig. \ref{fig:model}) can be stated as
\begin{equation}
{X_{i+1} = f(X_iA^*) .} 
\end{equation}
Here, $X_{i}$ and $X_{i+1}$ denote the input and output of $i^{\rm{th}}$ layer ($X_0$ means $F^{in}$, and $f$ is the learnable function used to change the dimensions of the feature vector of each node. 

\paragraph{Learning in LGA} The core of LGA relies on constructing a well representative directed graph.
The learning in this graph is guided by a direct loss $\mathcal{L}_{\rm LGA}$, on the output feature map, $F^{\rm out}$.
The main goal of LGA is to capture neighbouring context via message passing such that nodes corresponding to same information have similar distribution. Each node in the graph belongs to a particular spatial location in the image.
We assume that each node corresponds to a particular patch in the GT, $X^{\rm GT}$. This patch is calculated either via receptive field estimation of each node or by dividing the $X^{\rm GT}$ in $N$ equal patches where every node corresponds to a unique patch. Then we calculate the similarity between these patches. For example, for segmentation task, we calculate whether the patches have same classes. For image restoration tasks such as image dehazing, we calculate structural similarity index measure (SSIM) between the patches and if SSIM is greater than a predefined threshold, the patches are similar else not.

The nodes corresponding to similar patches are likely to have similar distribution. $\mathcal{L}_{\rm LGA}$ penalises $F^{\rm out}$ if nodes belonging to similar patches have different distribution or nodes corresponding to non-similar patches have similar distribution. This loss is defined such that LGA learns to predict richer representation to avoid learning an identity mapping between $F^{\rm in}$ and $F^{\rm out}$.

\paragraph{LGA contrastive loss} This novel loss term helps in the learning of our LGA module. Mathematically, LGA contrastive loss can be stated as

\begin{equation} \label{eq:divloss}
	{
		\begin{split}
			\mathcal{L}_{\rm LGA} = E_{P_i,P_j} \Bigg( \mathcal{C}_{ij}\log \left( \frac{V^{2}_{ij}}{U^{}_{ij}}  + 1\right)
			+ \bar{\mathcal{C}}_{ij}\log \left(
			\frac{U^{}_{ij}}{V^{2}_{ij}}  + 1\right)
			\Bigg)
		\end{split}
	}
\end{equation}

$V_{ij} = D(F^{\rm out}_i,F^{\rm out}_j)$ and $U_{ij} = D(F^{\rm in}_i,F^{\rm in}_j)$, $F^{\rm in}$ and $F^{\rm out}$ denote the input and output feature maps for the LGA module, and $D(\cdot)$ is the divergence function such as KL-divergence or mean square error. The two variables $\mathcal{C}$ and $\bar{\mathcal C}$ are boolean, and their values depend on the similarity between the neighboring nodes in the graph. For example, for nodes $i$ and $j$, we look at the GT labels of patches containing them. Further, we aggregate the labels to assign a single label to the node. For example, one aggregate measure could be to assign the majority class label to the node. Let $Agg(\cdot)$ denote the aggregate function and $i$ and $j$ denote two nodes from the graph, then $\mathcal{C} = 1$ if $Agg(i) = Agg(j)$, and 0 otherwise. 

Next, we present the interpretation and consequence of using this loss function. We note that $U_{ij}$ is determined by the input to the LGA module and does not update as LGA learns. LGA essentially learns through tweaking the value of $V_{ij}$, which is determined by the output of LGA. Note that larger diverse value $D_{ij}$ indicates larger difference in the distributions of the nodes $i$ and $j$. With these points, we now assess how the loss function behaves and influences the learning of LGA in different situations. For the case when two nodes have a similar distribution, $V_{ij}$ needs to be minimized, implying that the two nodes have similar output distributions as well. For cases where the input distributions of the two nodes are very different, $V_{ij}$ is learnt to be maximized thereby setting the two output distributions apart. However, using only $V_{ij}$ is not useful since the encoder is also learning to generate better input feature maps. Even if the LGA does not improve, the feature map will give $V_{ij}$ equivalent to $U_{ij}$, the overall loss will decrease, and it would look as if LGA is learning. Hence we also incorporated $U_{ij}$ in the loss function. More details can be found in Appendix \ref{sec:LGA contrastive loss analysis} of this paper.

\subsection{Other features of LGA}\label{sec:efficiency}

It has been discussed in Sec. \ref{sec:LGA} how the loss function of LGA is designed to not only help the LGA learn, but also to influence the learning in the preceding or succeeding networks. It has also been discussed in Sec. \ref{subsec:training process} how the LGA can be easily incorporated in different architectures in a modular fashion. 
There are two other major benefits of LGA, namely efficiency and scalability, in comparison to globally connected graph and attention models. 

\paragraph{Efficiency} Constructing a globally connected graph \emph{i.e.} a graph where any node can be connected to any other node is a computationally demanding task. For our LGAs, we need only  $9 \times N$ different $1 \times 1$ convolutional layers, where $N$ is the number of nodes. This solution is more space and time efficient than most other approaches. The LGA needs only $\mathcal{O}(N)$ space for storing edge weights, whereas the spatial complexity is mostly of the order $\mathcal{O}(N^2)$ in attention models or globally connected graph. CCNet attention \cite{huang2020ccnet} has a spatial complexity of $\mathcal{O}(N^{1.5})$, which is still higher than our model. With respect to time complexity, the usual attention models or globally connected graph have a complexity of the order $\mathcal{O}(NC^2 + N^2C)$, here $NC^2$ comes from applying convolution on feature map for channel reduction and $N^2C$ comes from the information propagation. The term $NC^2$ can be reduced by a significant amount by applying depth-wise or group convolution. Hence, the best optimized complexity for such models becomes $\mathcal{O}(N^2C)$. In the case of CCNet, the complexity is  $\mathcal{O}(NC^2 + N^{1.5}C)$ and the optimized complexity is $\mathcal{O}(N^{1.5}C)$. On the other hand for our network, the optimized time complexity is $\mathcal{O}(NC)$ because our module's layer takes only $\mathcal{O}(NC)$ for information propagation. As there are 4 layers in our LGA, the complexity is $\mathcal{O}(4NC)$. Which is $\sqrt{N}$ times smaller than the best light-weight attention model (CCNet). Further, we have provide the derivation of the space time complexity of LGA vs. CCNet in Appendix \ref{sec:complexity analysis}.

\paragraph{Scalability} Here, we use $1 \times 1$ convolutional layer assuming that a local edge (i.e edge between spatially close nodes) can be constructed using the information of the node itself. However, incorporating information of a larger context is possible simply by using convolutional layers of larger sizes, for example, kernel of size $2 \times 2$ or higher. Similarly, although we have currently used only 4 layers of LGA, this can be easily extended to have more layers such that a larger scale of global context can be included. In comparison, it is difficult to incorporate scalability and correlate features that are spatially far apart in attention models.

\paragraph{Effective scene context capture} As our LGA module propagates information between two nodes by passing information via in-between neighbouring nodes, it gradually captures the context of surrounding nodes i.e non-local context for information propagation. Also, our LGA module automatically takes distance between nodes into consideration, since information from a node reaches its neighbour earlier and in greater amount effectively, thereby implicitly embedding that the nodes closer to each other are more likely to have a similar structure when LGA is included.

\begin{figure*}[t]
\centering
\scalebox{1}{
\begin{tabular}{cccccccc}
    \footnotesize{INPUT} & \footnotesize{ESNet} & \footnotesize{LEDNet} & \footnotesize{ViT} & \footnotesize{PSPNet} & \footnotesize{DeepLabV3} & \footnotesize{Squeeze} & \footnotesize{Ground}\\[-4pt]
    & & & & & & \footnotesize{with LGA} & \footnotesize{truth}\\
    \includegraphics[width=0.1 \linewidth,height=0.1 \linewidth]{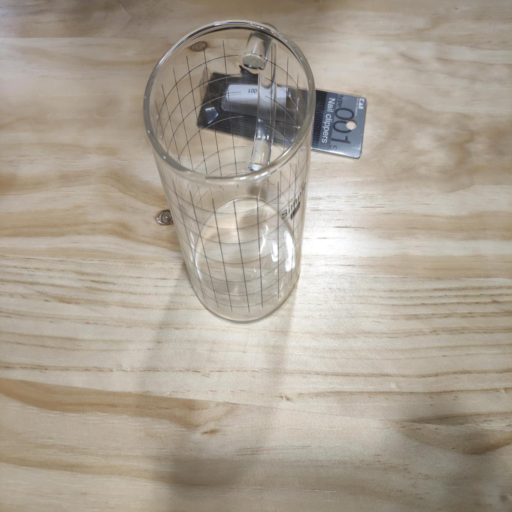} & 
    \includegraphics[width=0.1 \linewidth,height=0.1 \linewidth]{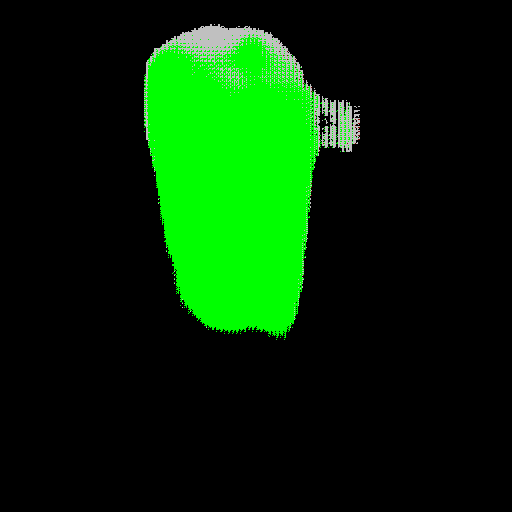} &
    \includegraphics[width=0.1 \linewidth,height=0.1 \linewidth]{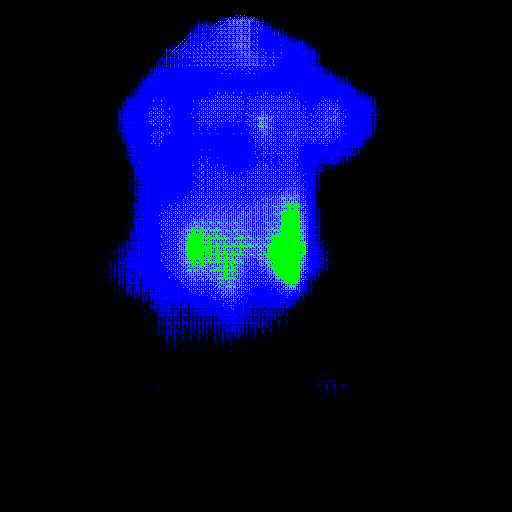} & 
    \includegraphics[width=0.1 \linewidth,height=0.1 \linewidth]{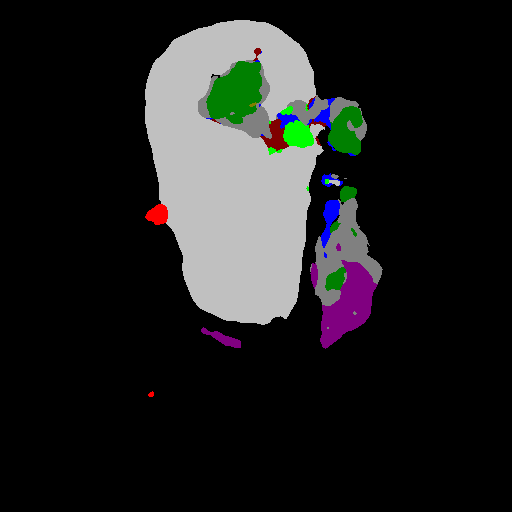} &  
    \includegraphics[width=0.1 \linewidth,height=0.1 \linewidth]{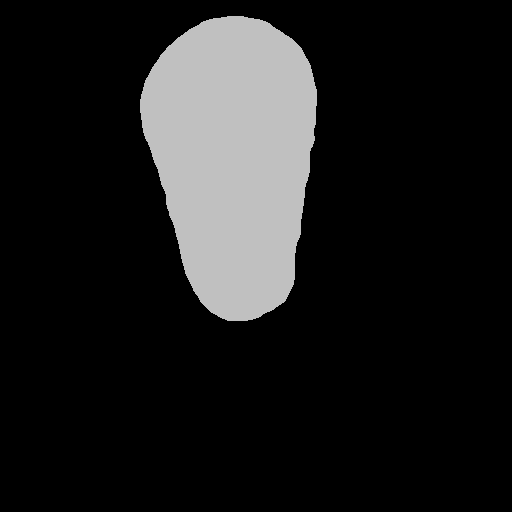} &
    \includegraphics[width=0.1 \linewidth,height=0.1 \linewidth]{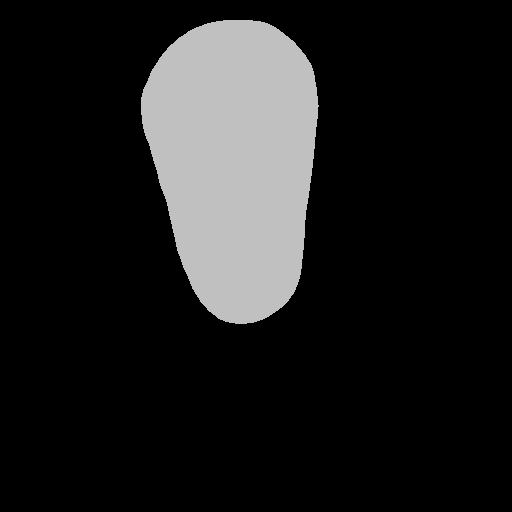} &
    \includegraphics[width=0.1 \linewidth,height=0.1 \linewidth]{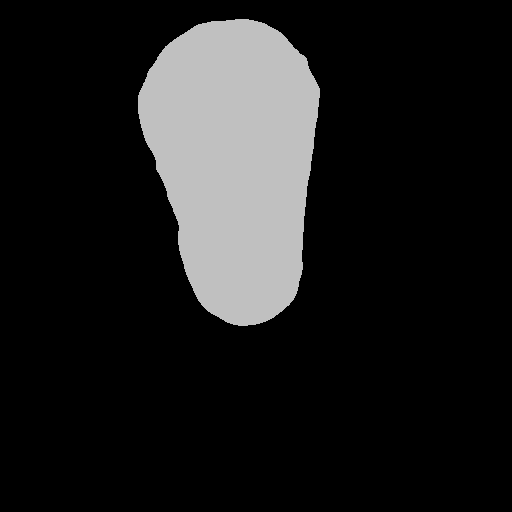} & 
    \includegraphics[width=0.1 \linewidth,height=0.1 \linewidth]{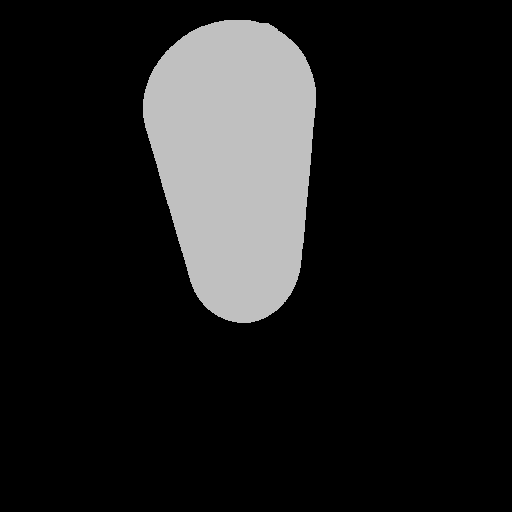} \\
    \includegraphics[width=0.1 \linewidth,height=0.1 \linewidth]{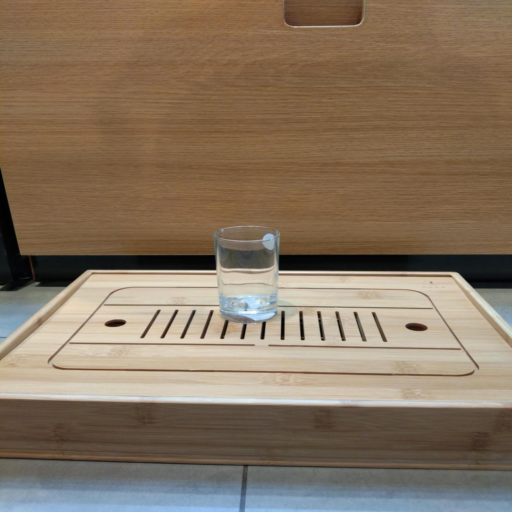} &  
    \includegraphics[width=0.1 \linewidth,height=0.1 \linewidth]{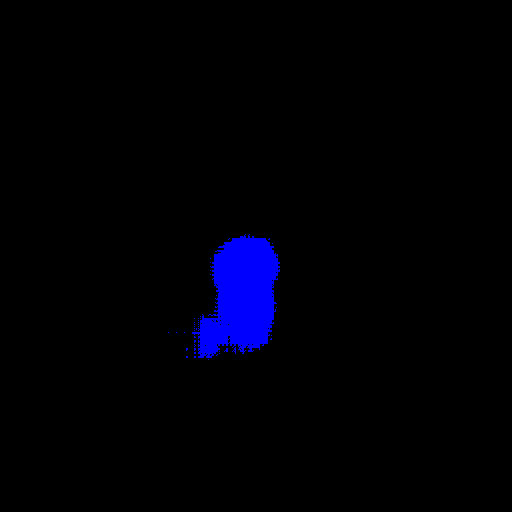} &
    \includegraphics[width=0.1 \linewidth,height=0.1 \linewidth]{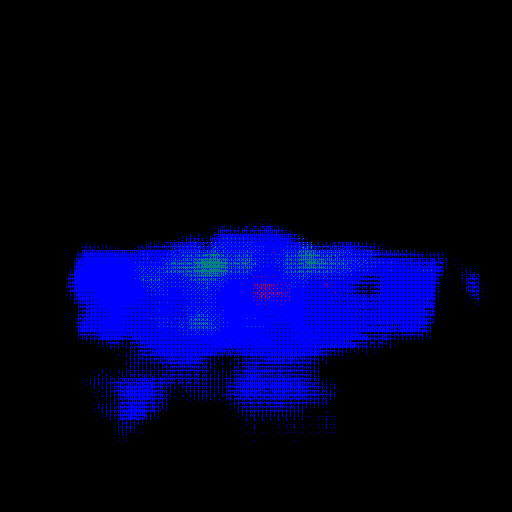} & 
    \includegraphics[width=0.1 \linewidth,height=0.1 \linewidth]{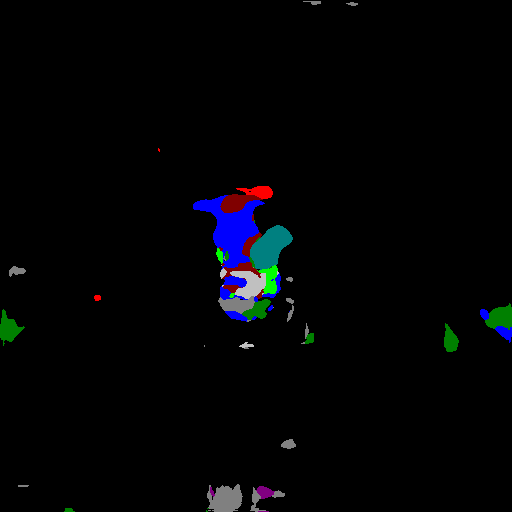} &  
    \includegraphics[width=0.1 \linewidth,height=0.1 \linewidth]{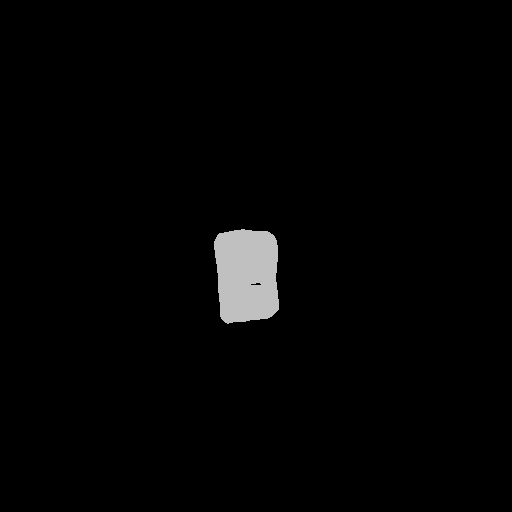} &
    \includegraphics[width=0.1 \linewidth,height=0.1 \linewidth]{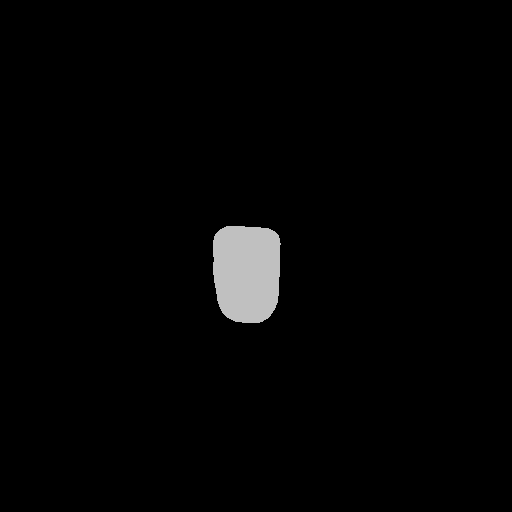} &
    \includegraphics[width=0.1 \linewidth,height=0.1 \linewidth]{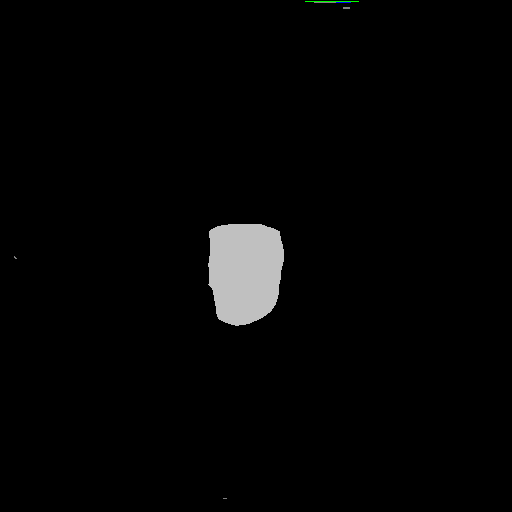} & 
    \includegraphics[width=0.1 \linewidth,height=0.1 \linewidth]{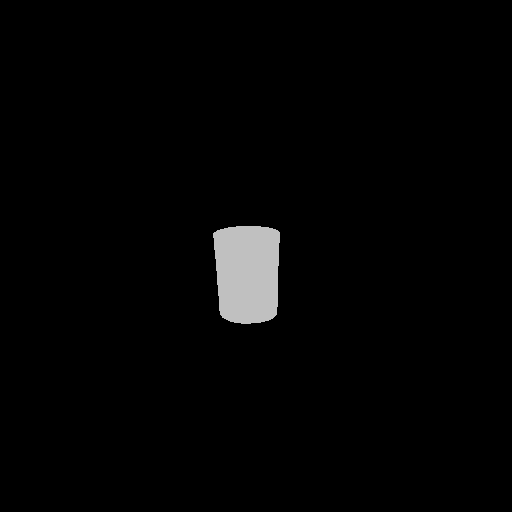} \\
    \includegraphics[width=0.1 \linewidth,height=0.1 \linewidth]{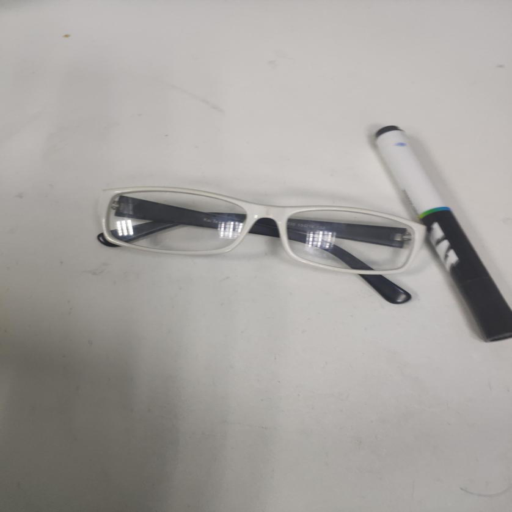} &  
    \includegraphics[width=0.1 \linewidth,height=0.1 \linewidth]{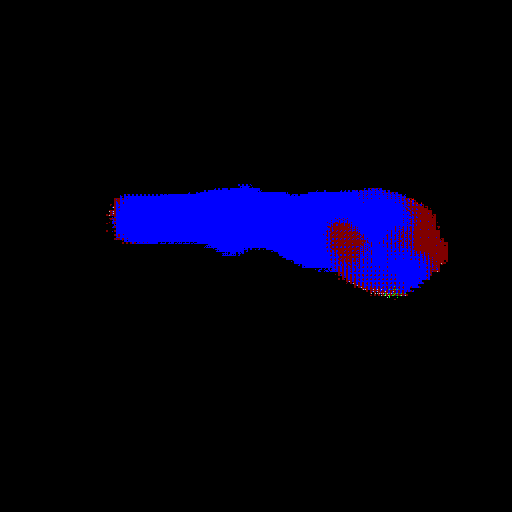} &
    \includegraphics[width=0.1 \linewidth,height=0.1 \linewidth]{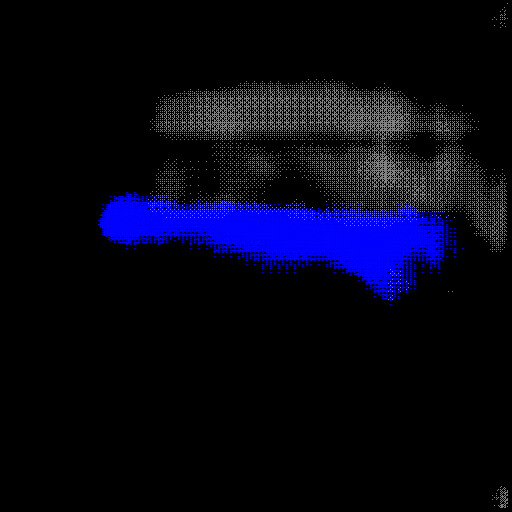} & 
    \includegraphics[width=0.1 \linewidth,height=0.1 \linewidth]{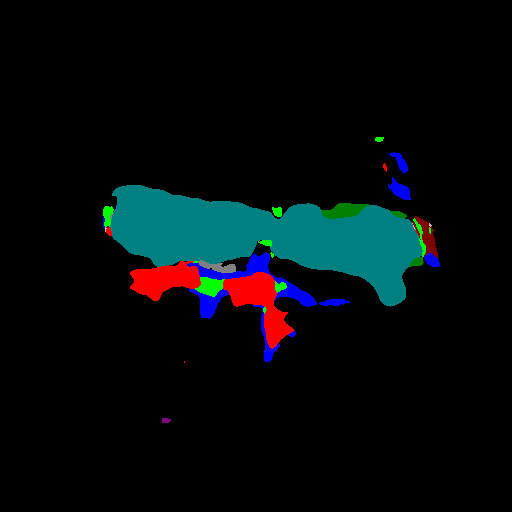} &  
    \includegraphics[width=0.1 \linewidth,height=0.1 \linewidth]{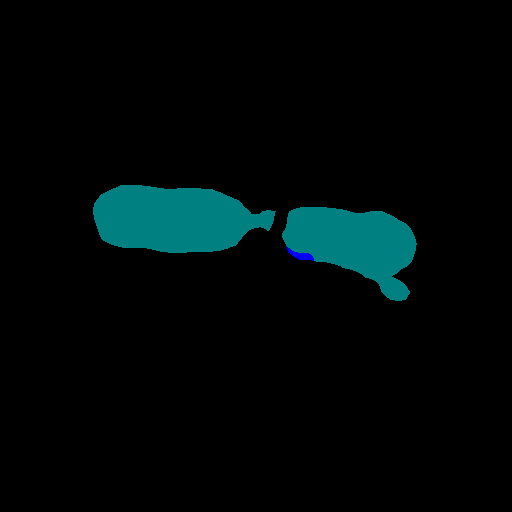} &
    \includegraphics[width=0.1 \linewidth,height=0.1 \linewidth]{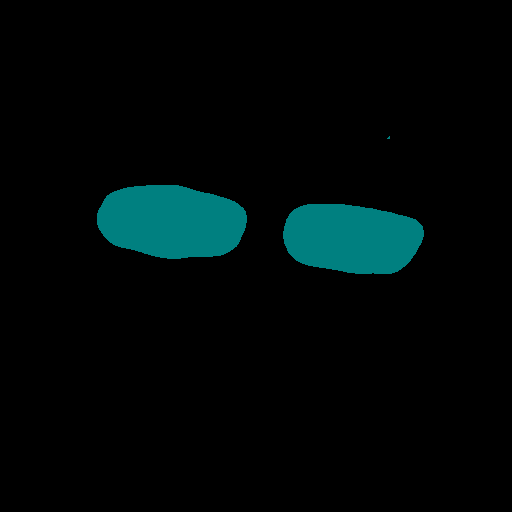} &
    \includegraphics[width=0.1 \linewidth,height=0.1 \linewidth]{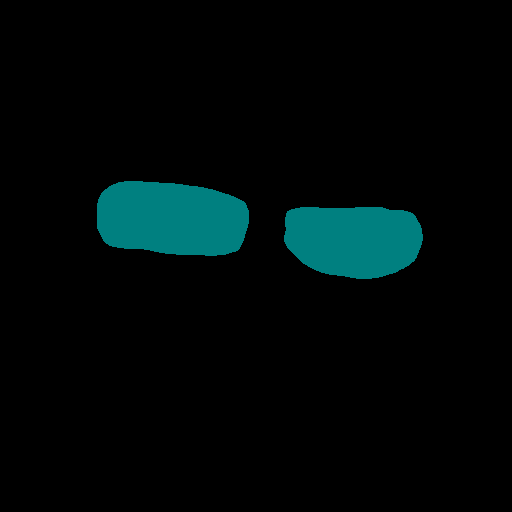} & 
    \includegraphics[width=0.1 \linewidth,height=0.1 \linewidth]{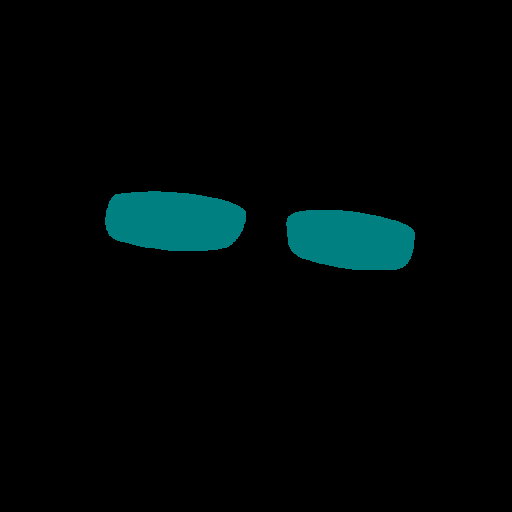} \\
\end{tabular}
}
\caption{Examples of segmentation results of previous methods are presented. Eleven classes of objects are color coded. }
\label{fig:segmentation}
\end{figure*}

\section{Experiments}

\subsection{Training process}\label{subsec:training process}
We converted all the input images and ground truth for all segmentation into shape $512 \times 512$. The optimizer used in our training process was Adam~\cite{kingma2014adam}. For segmentation, we use various encoders such as squeezenet, etc and add our own decoder network. We added the LGA layer after the encoder and passed encoder output as the LGA input. For  segmentation experiments, an initial learning rate of $10^{-4}$ is used, and after 30 epochs, we reduce it to $10^{-5}$. We train the models for 40 epochs. For dehazing, we used ~\cite{singh2020single} as the base method. We reduced its size and FLOPs by replacing the standard convolutional layer with a group convolution layer. It is represented as BPPNet-reduced in this work. LGA is added after the stacked UNet layer in ~\cite{singh2020single}. The choice of learning rate and the decay mechanism are chosen to be similar to that of training a generator as described in ~\cite{singh2020single}. The dataset used for segmentation is Trans10Kv2~\cite{translab2}. This dataset is the modification of Trans10k ~\cite{translab2} with more fine-tuned classes \emph{i.e.} 12. For dehazing, we used I-Haze dataset ~\cite{DBLP:journals/corr/abs-1804-05091}. For optical flow estimation, we used ARFlow \cite{liu2020learning} architecture as the base method. We have added LGA layer after the feature extraction layer. We only applied LGA at a single feature map of spatial size $28 \times 24$. Additional details can be found in Appendix \ref{sec:architectural details} and Appendix \ref{sec:incorp. of LGA} of this paper.

\begin{table}[t]
	\setlength{\tabcolsep}{03pt}	
	\renewcommand{\arraystretch}{1}
	\caption{Segmentation results: quantitative comparison of various models.}
	\centering
	\resizebox{!}{!}{
		\begin{tabular}{lcccc}
			\toprule
			\multirow{2}{*}{MODEL} & \multicolumn{2}{c}{Performance ($\uparrow$)}                                                                               & \multicolumn{2}{c}{Efficiency ($\downarrow$)}                                                                                                                           \\ \cmidrule(r){2-3} \cmidrule(l){4-5} 
			& \begin{tabular}[c]{@{}c@{}}mIoU\\ (\%)\end{tabular} & \begin{tabular}[c]{@{}c@{}}Accuracy\\ (\%)\end{tabular} & \begin{tabular}[c]{@{}c@{}}Parameters\\ ($\times 10^6$)\end{tabular} & \begin{tabular}[c]{@{}c@{}}FLOPS\\ ($\times 10^9$)\end{tabular} \\ \midrule
			\multicolumn{5}{c}{Efficient/real-time small architectures}                                                                                                                                                                                                                            \\ \midrule
			FPENet \cite{fpe}                 & 10.1                                                & 70.3                                                    & 0.5                                                                  & 0.8                                                       \\
			ESPNet-v2 \cite{mehta2019espnetv2}             & 12.3                                                & 73.0                                                    & 3.5                                                                           & 0.8                                                             \\
			ENet \cite{paszke2016enet}                  & 23.4                                                & 78.2                                     & 0.4                                                            & 2.1                                                                      \\
			DABNet \cite{li2019dabnet}                 & 15.3                                                & 77.4                                                    & 0.8                                                                           & 5.2                                                                      \\
			LEDNet \cite{wang2019lednet}                 & 30.3                                       & 72.9                                                    & 1.1                                                                           & 19.6                                                                     \\
			ICNet \cite{zhao2018icnet}                 & 23.4                                                & 78.2                                                    & 7.8                                                                           & 10.6                                                                     \\
			MobileNet-v2 \cite{sandler2018mobilenetv2}          & 17.6                                                & 77.6                                           & 3.3                                                                           & 29.3                                                                     \\
			ESNet \cite{wang2019esnet}                 & 43.6                                 & 45.5                                                    & 1.7                                                                           & 27.3                                                                     \\ \midrule
			\multicolumn{5}{c}{LGA incorporation into small architectures}                                                                                                                                                                                                                                                    \\ \midrule
			Shuffle\cite{zhang2017shufflenet} with LGA                & 44.5                                                & 78.7                                                    & 0.4                                                                           & 3.5                                                                      \\
			Squeeze\cite{Squeeze1} with LGA               & 44.6                                                & 79.6                                                    & 1.1                                                                           & 13.5                                                                     \\ \midrule
			\multicolumn{5}{c}{Large architectures}                                                                                                                                                                                                                                          \\ \midrule
			ViT \cite{dosovitskiy2020image}                   & 29.6                                                & 67.8                                                    & 171.6                                                                         & 176.7                                                                    \\
			PSPNet (Res34) \cite{zhao2017pyramid}         & 43.2                                                & 82.8                                                    & 21.5                                                           & 19.3                                                      \\
			PSPNet (Res50) \cite{zhao2017pyramid}         & 43.2                                                & 83.2                                                    & 24.4                                                                 & 24.0                                                            \\
			DeepLab \cite{DeepLabv3}               & 59.1                                 & 89.6                                                    & 39.6                                                                          & 328.0                                                                    \\ \bottomrule
		\end{tabular}
	}
	\label{tab:segmentation}
\end{table}

\subsection{Segmentation }
We compare LGA-incorporated Squeeze and Shuffle with previous challenging methods. Both of the base models can operate in real-time and are computationally efficient. however, there performance is not at part with the bulkier and/or computationally more intensive models (see Table \ref{tab:segmentation} and Fig. \ref{fig:segmentation}).
We see that models that include LGA outperform all the efficient and real-time architectures in terms of performance. Shuffle with LGA has the smallest number of parameters as well. Further both our models are computationally less expensive, but the performance is at par with the lager models.
In terms of comparison with larger models that comprise millions of parameters and need billions of FLOPs, our models are less demanding, Also, these perform better than PSPNet in terms of mIoU, although marginally inferior in terms of SSIM. We see that DeepLab performs the best, however this model is relatively very large. Nonetheless, it is evident that for real-time or compact application scenarios, LGA provides an efficient and performance-effective solution.

\begin{figure*}[t]
\centering
\scalebox{1}{
\begin{tabular}{cccccccccc}
\footnotesize{INPUT} & \footnotesize{He} & \footnotesize{Zhu} & \footnotesize{Ren} & \footnotesize{Berman} & \footnotesize{Li} & \footnotesize{BPP} & \footnotesize{BPP} & \footnotesize{BPP} & \footnotesize{Ground} \\[-4pt]
& \footnotesize{\emph{et al.} \cite{he2010single}} & \footnotesize{\emph{et al.} \cite{zhu2015fast}} & \footnotesize{\emph{et al.} \cite{ren2016single}} & \footnotesize{\emph{et al.} \cite{berman2016non}} & \footnotesize{\emph{et al.} \cite{li2017all}} & \footnotesize{reduced} & & \footnotesize{with LGA} & \footnotesize{Truth}\\

    \includegraphics[width=0.077 \linewidth,height=0.077 \linewidth]{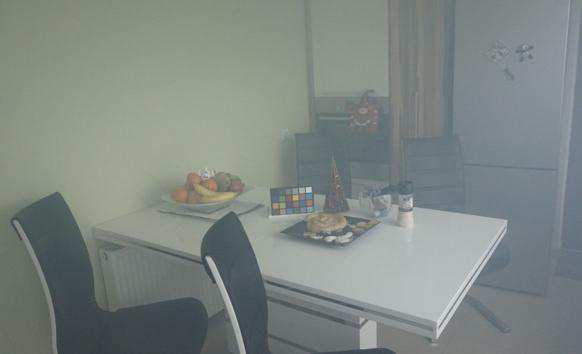} &  
    \includegraphics[width=0.077 \linewidth,height=0.077 \linewidth]{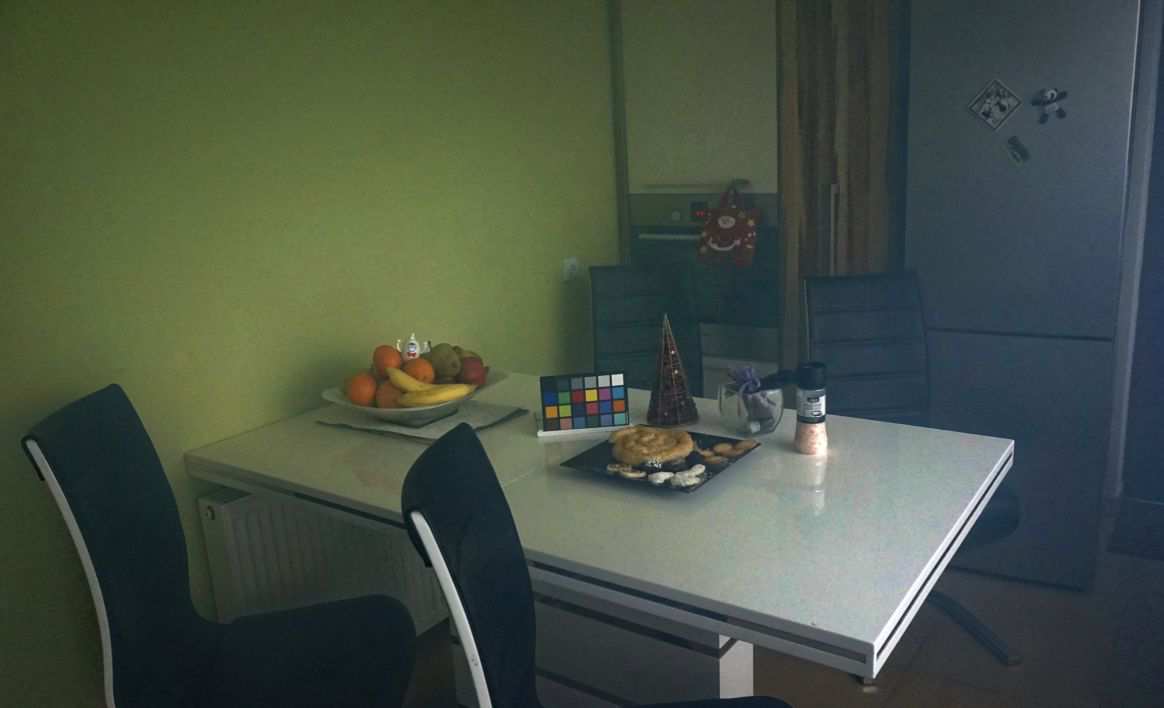} &
    \includegraphics[width=0.077 \linewidth,height=0.077 \linewidth]{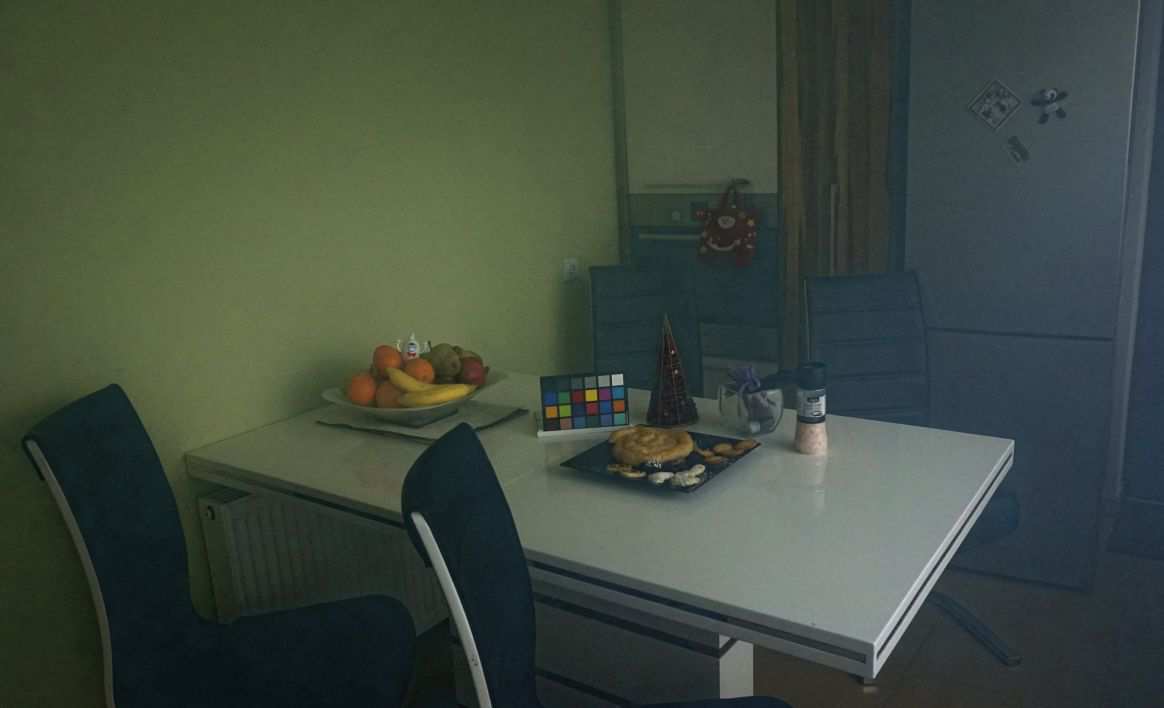} &  
    \includegraphics[width=0.077 \linewidth,height=0.077 \linewidth]{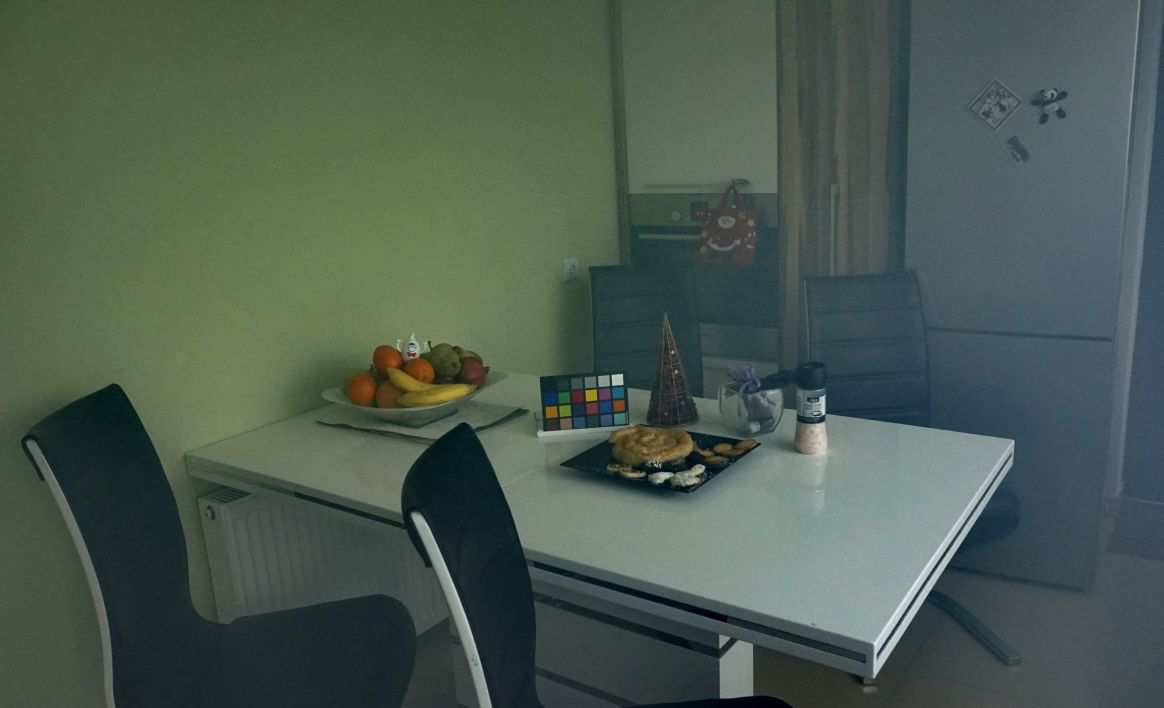} &
    \includegraphics[width=0.077 \linewidth,height=0.077 \linewidth]{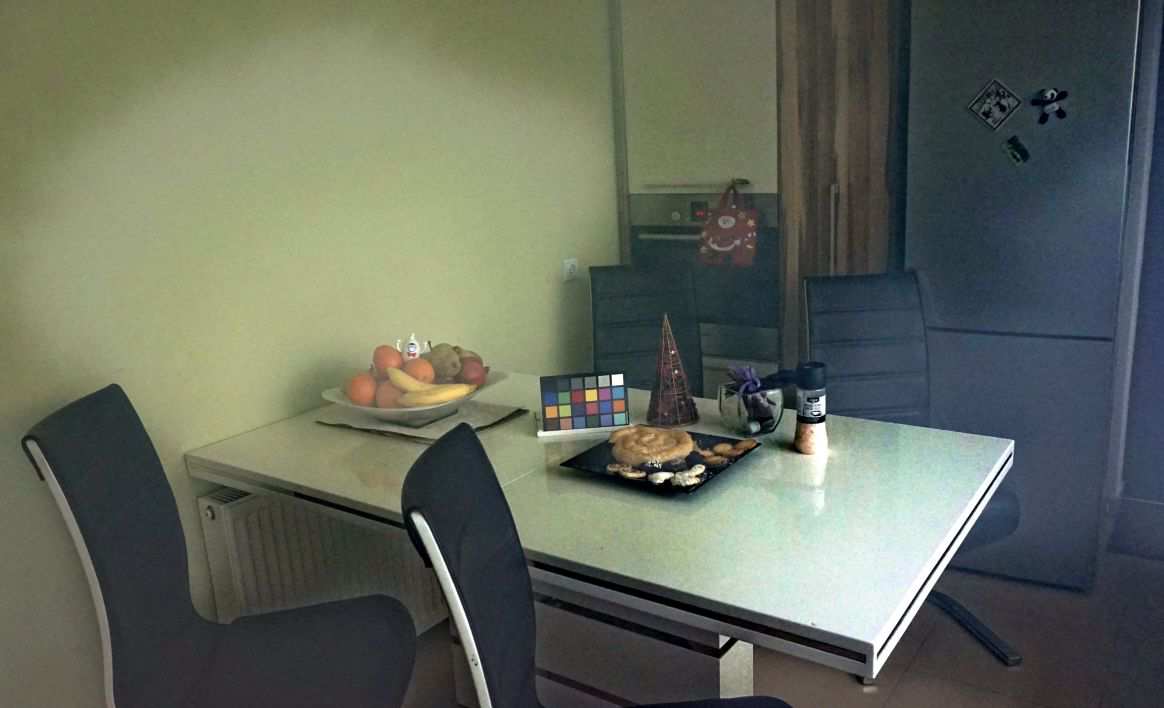} &  
    \includegraphics[width=0.077 \linewidth,height=0.077 \linewidth]{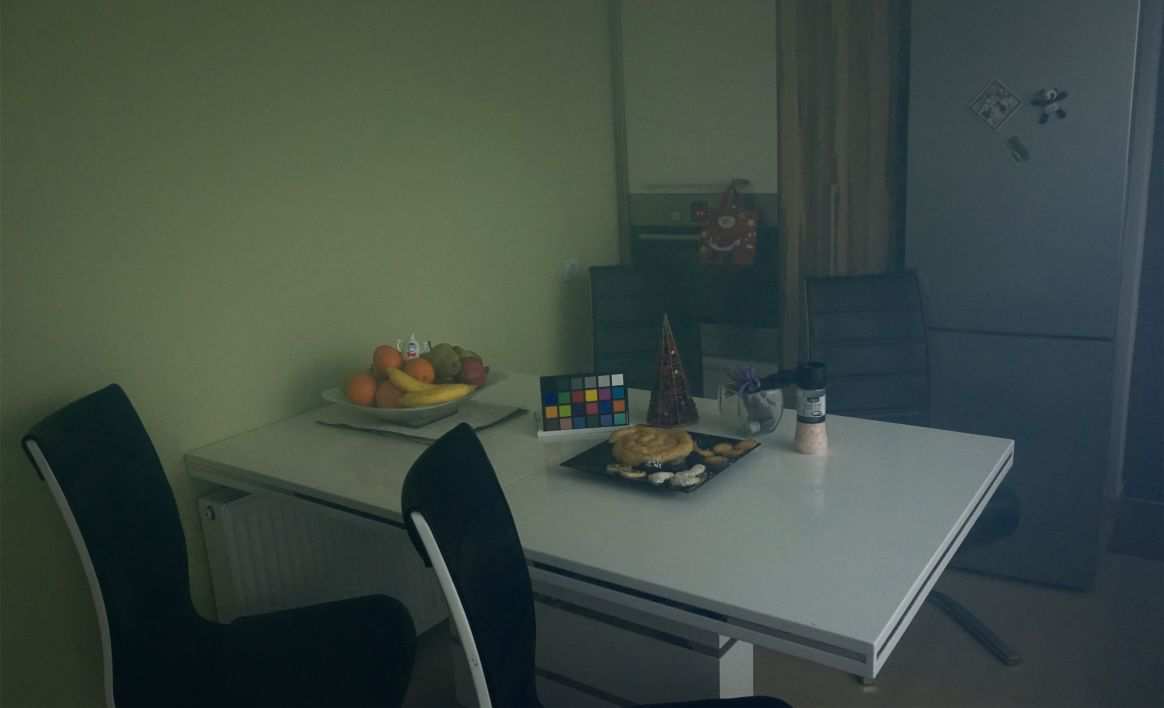} &
    \includegraphics[width=0.077 \linewidth,height=0.077 \linewidth]{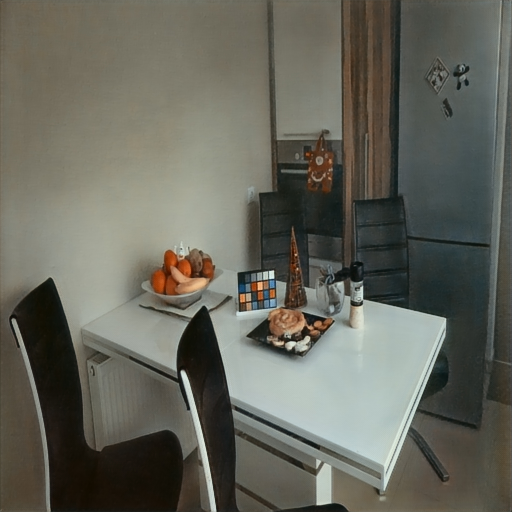} &  
    \includegraphics[width=0.077 \linewidth,height=0.077 \linewidth]{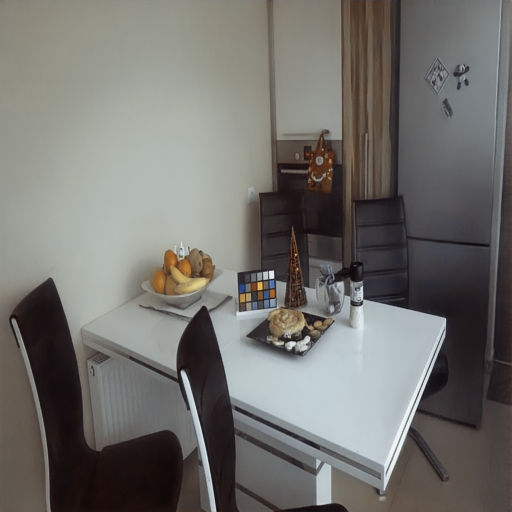} &
    \includegraphics[width=0.077 \linewidth,height=0.077 \linewidth]{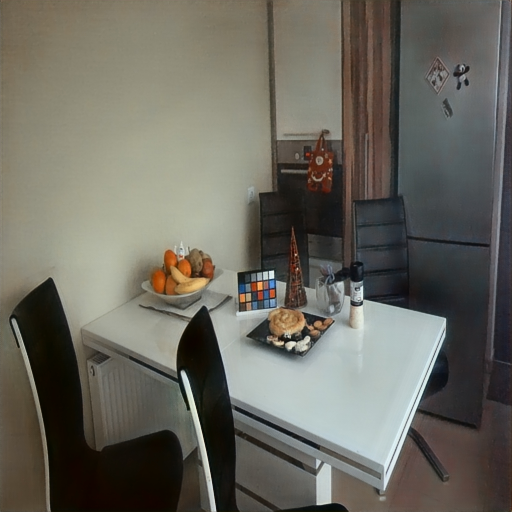} &
    \includegraphics[width=0.077 \linewidth,height=0.077 \linewidth]{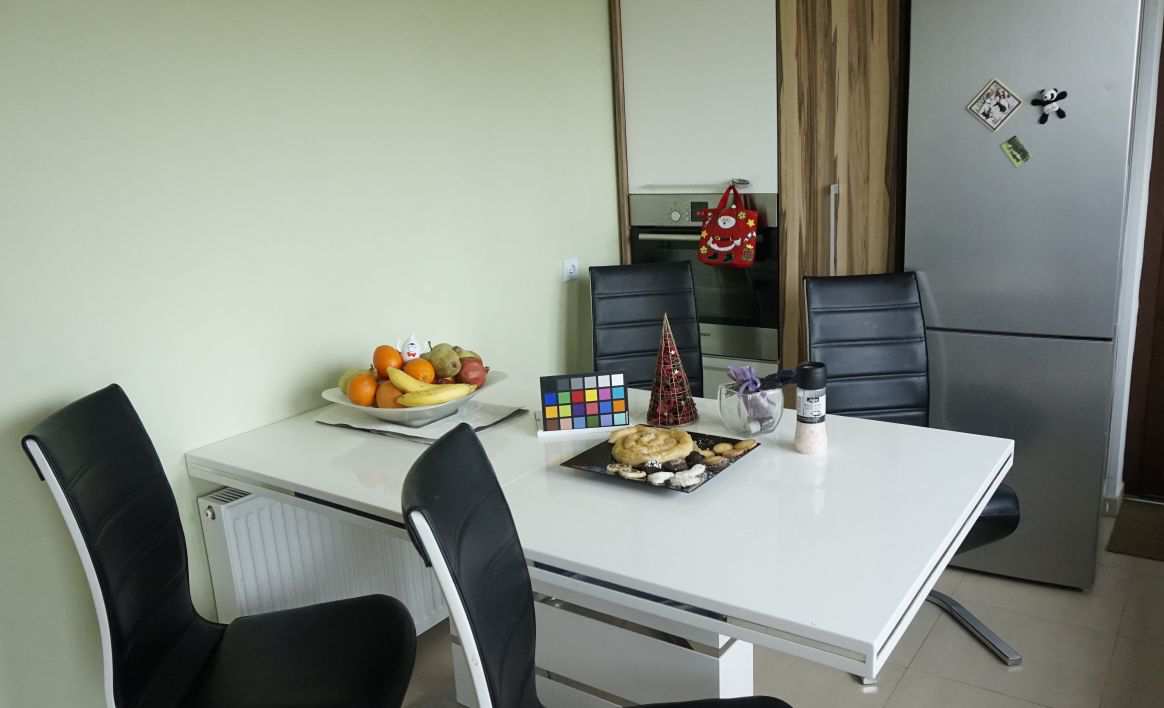} \\
    \includegraphics[width=0.077 \linewidth,height=0.077 \linewidth]{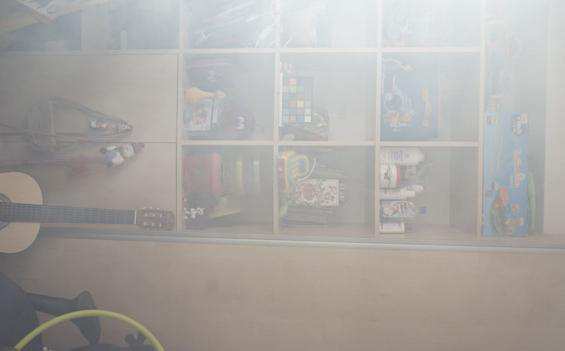} &  
    \includegraphics[width=0.077 \linewidth,height=0.077 \linewidth]{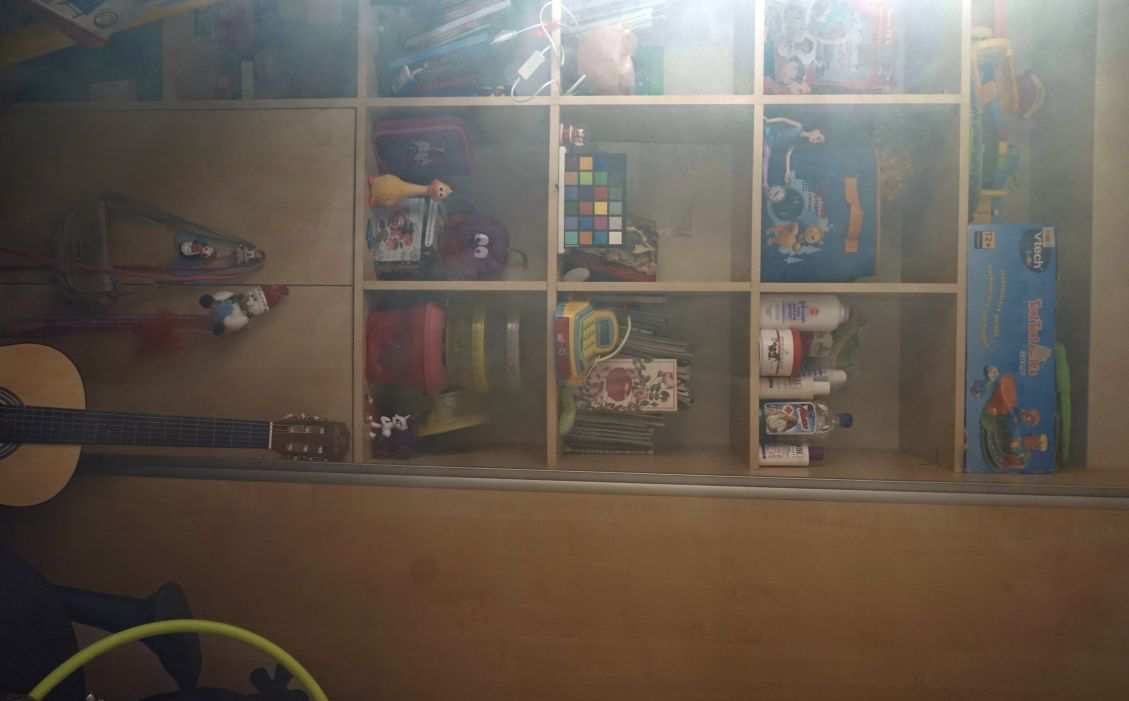} &
    \includegraphics[width=0.077 \linewidth,height=0.077 \linewidth]{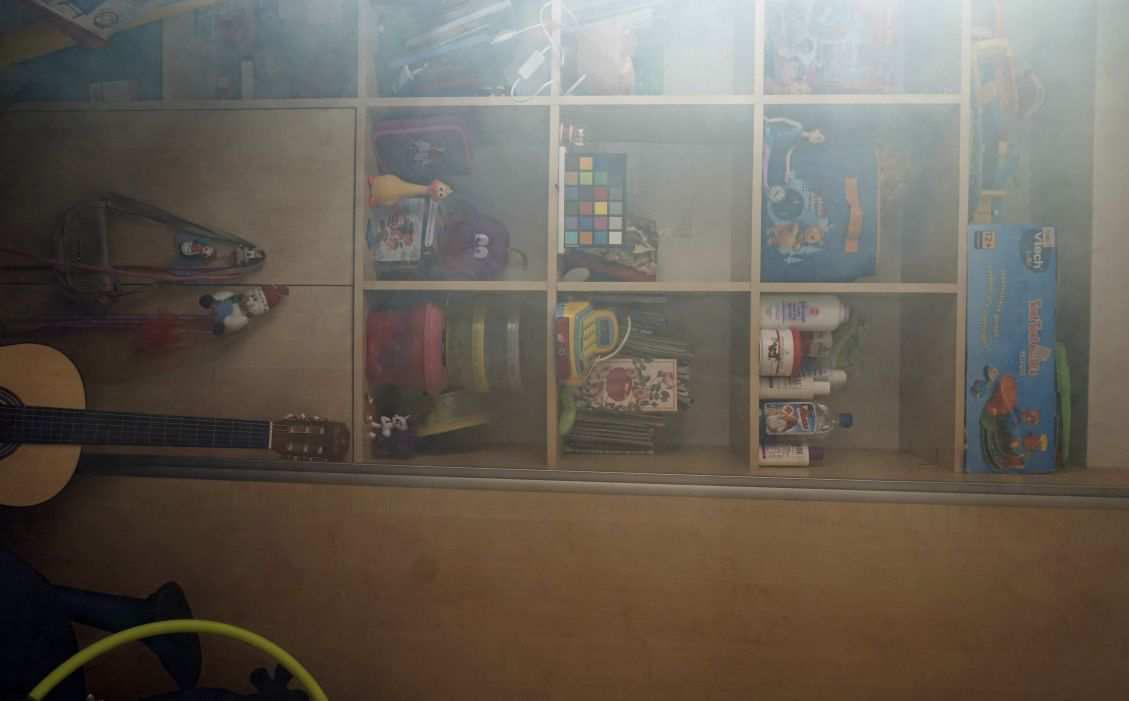} &  
    \includegraphics[width=0.077 \linewidth,height=0.077 \linewidth]{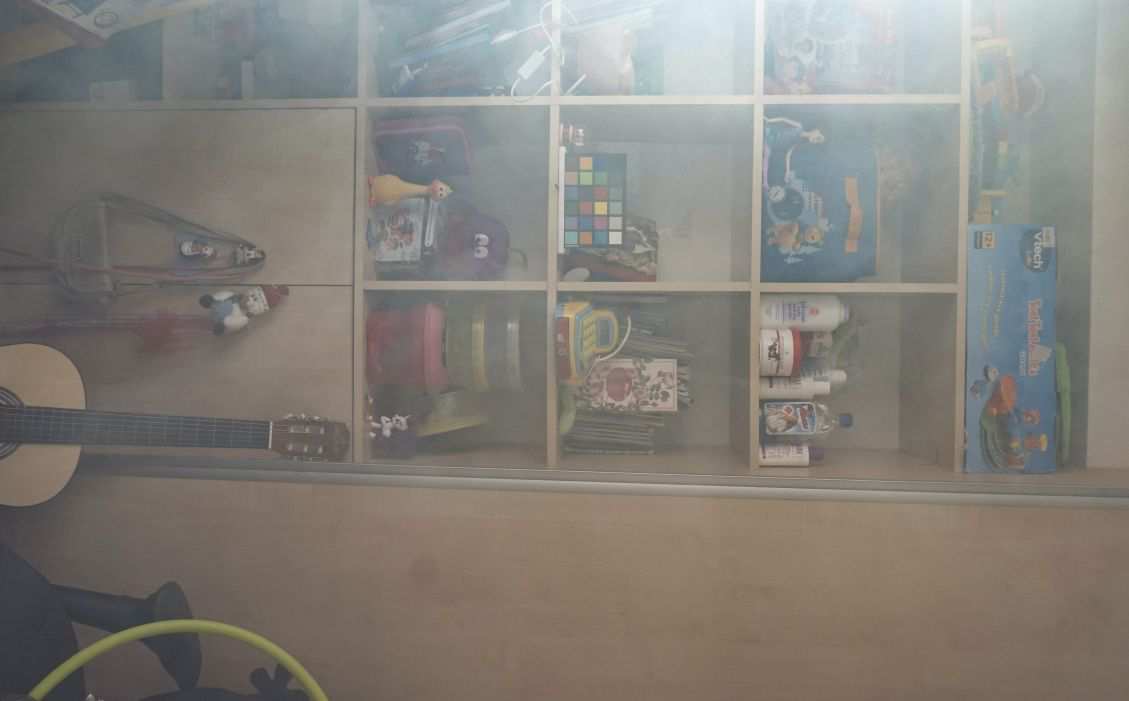} &
    \includegraphics[width=0.077 \linewidth,height=0.077 \linewidth]{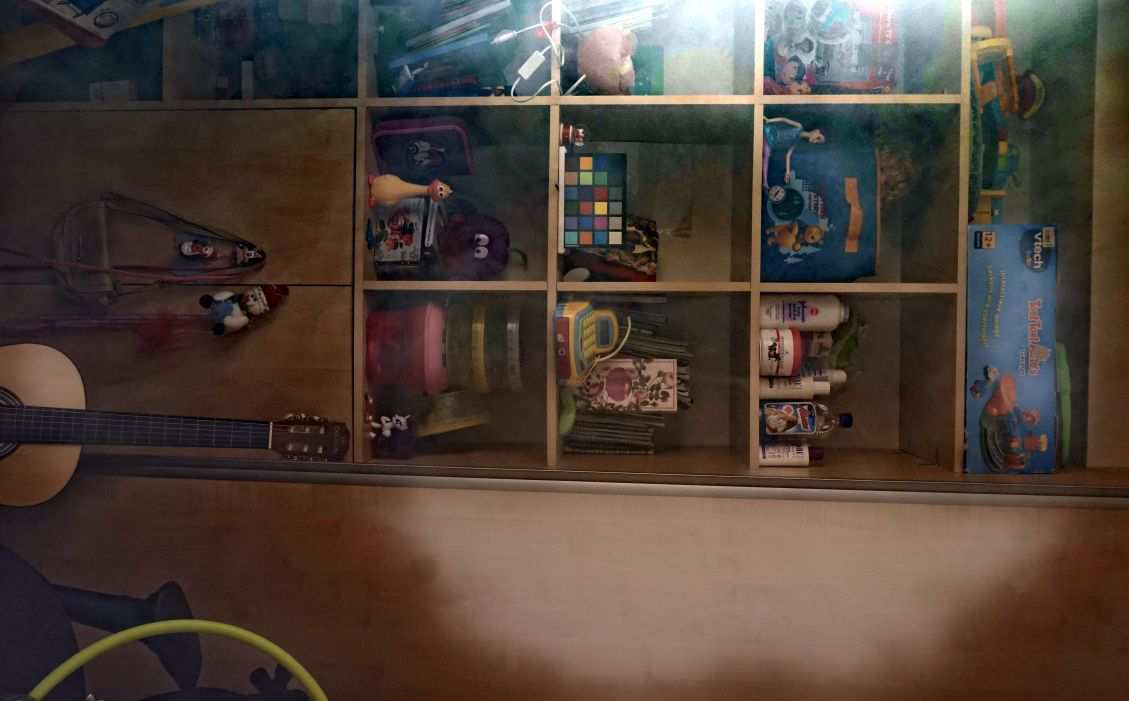} &  
    \includegraphics[width=0.077 \linewidth,height=0.077 \linewidth]{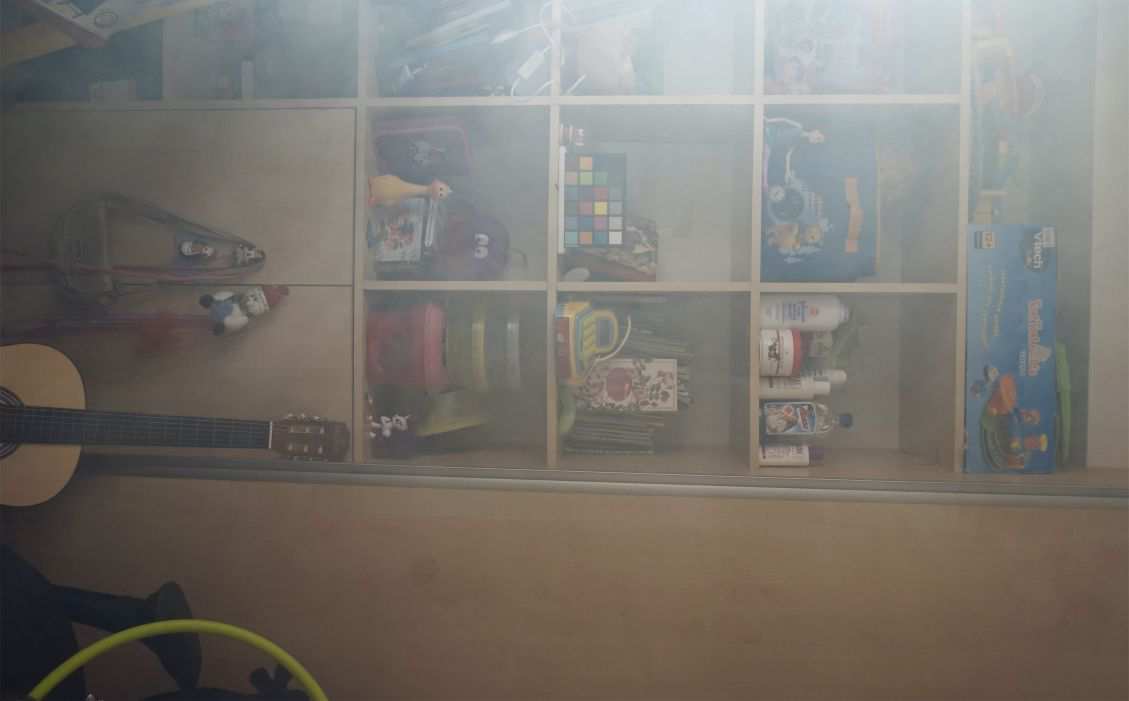} &
    \includegraphics[width=0.077 \linewidth,height=0.077 \linewidth]{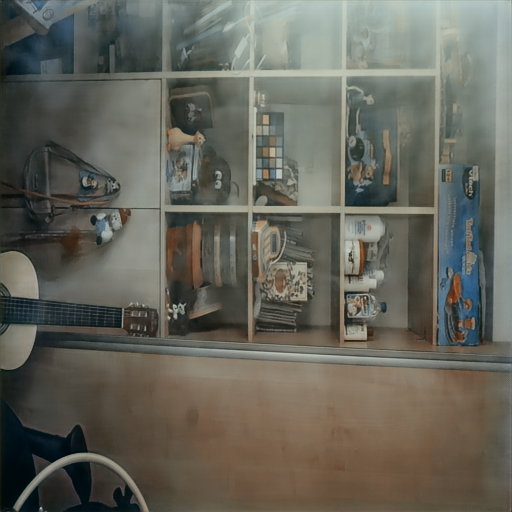} &  
    \includegraphics[width=0.077 \linewidth,height=0.077 \linewidth]{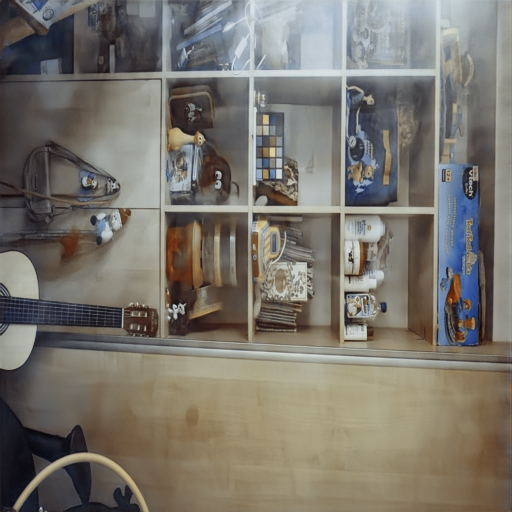} &
    \includegraphics[width=0.077 \linewidth,height=0.077 \linewidth]{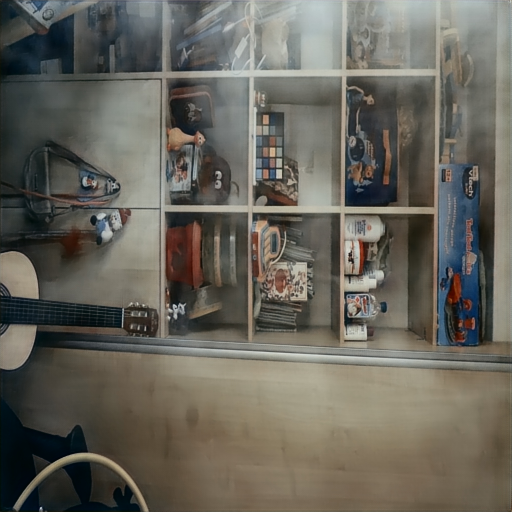} &
    \includegraphics[width=0.077 \linewidth,height=0.077 \linewidth]{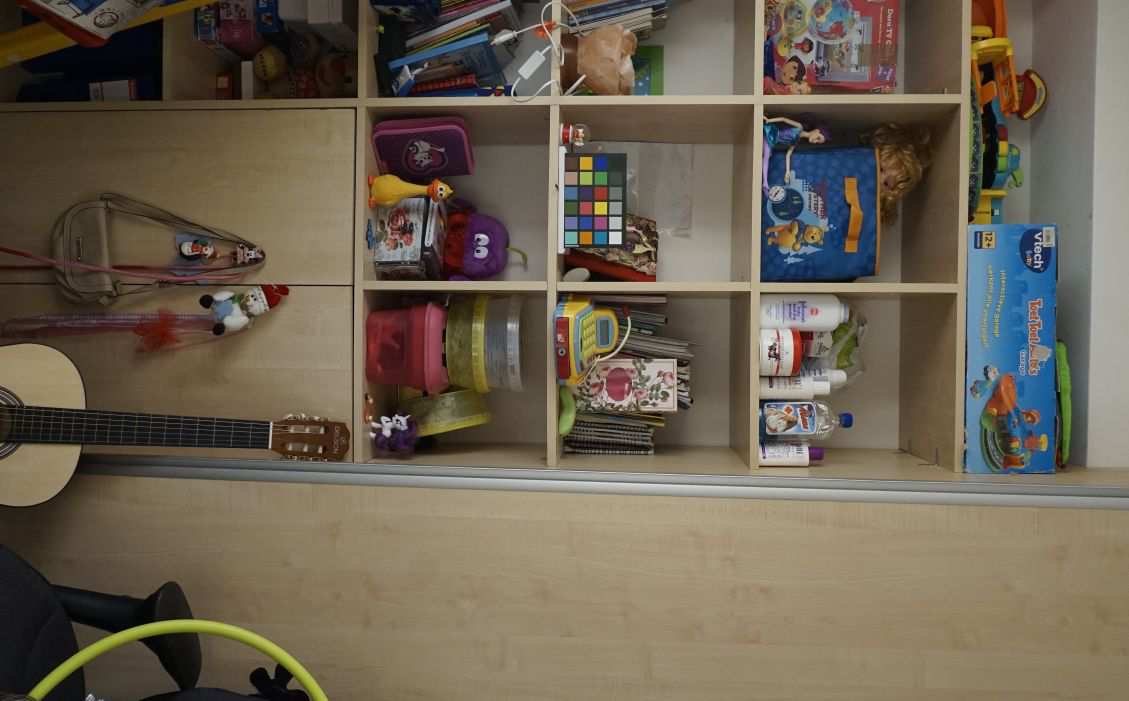} \\
    \includegraphics[width=0.077 \linewidth,height=0.077 \linewidth]{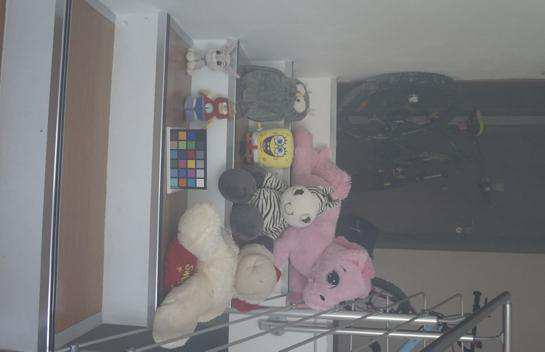} &  
    \includegraphics[width=0.077 \linewidth,height=0.077 \linewidth]{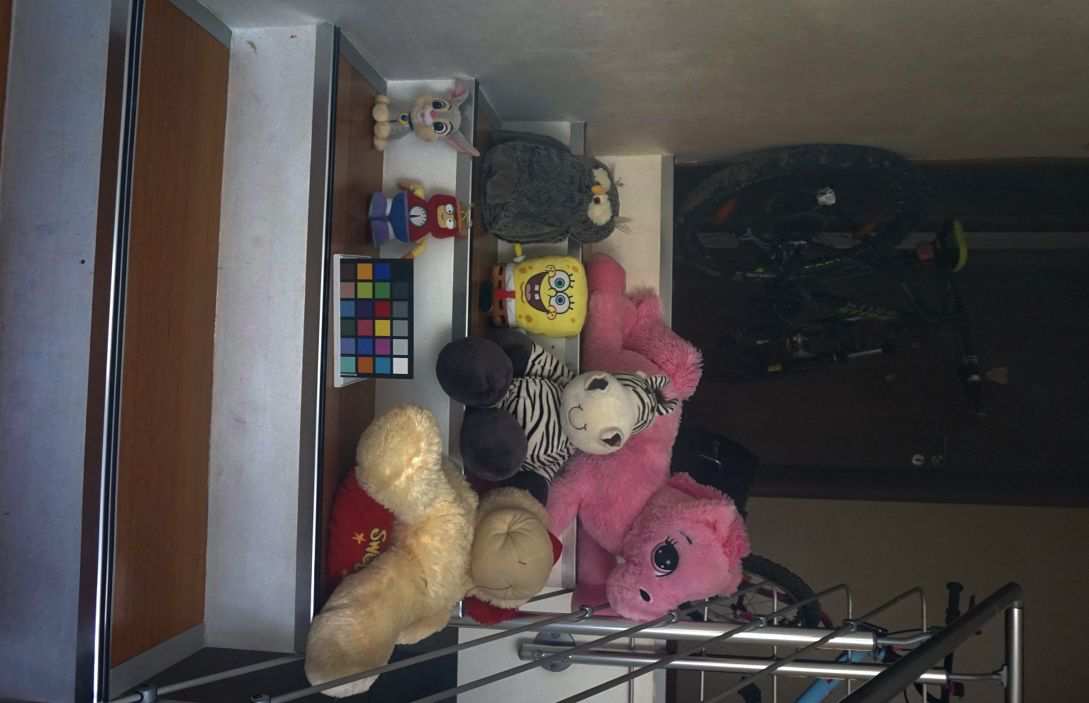} &
    \includegraphics[width=0.077 \linewidth,height=0.077 \linewidth]{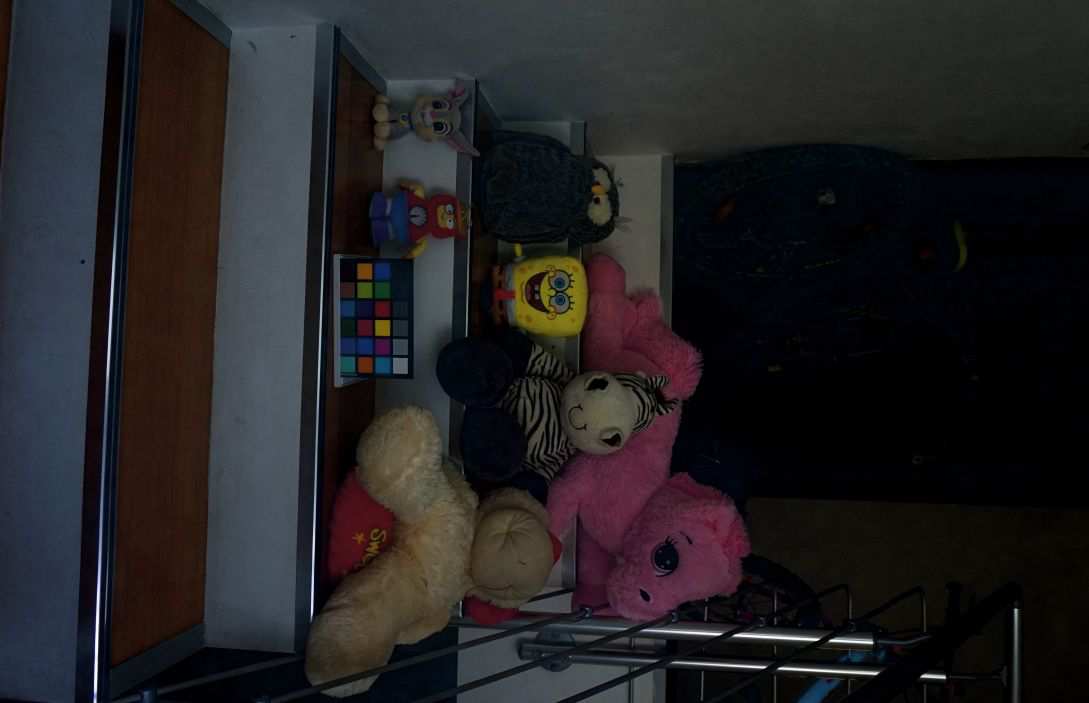} &  
    \includegraphics[width=0.077 \linewidth,height=0.077 \linewidth]{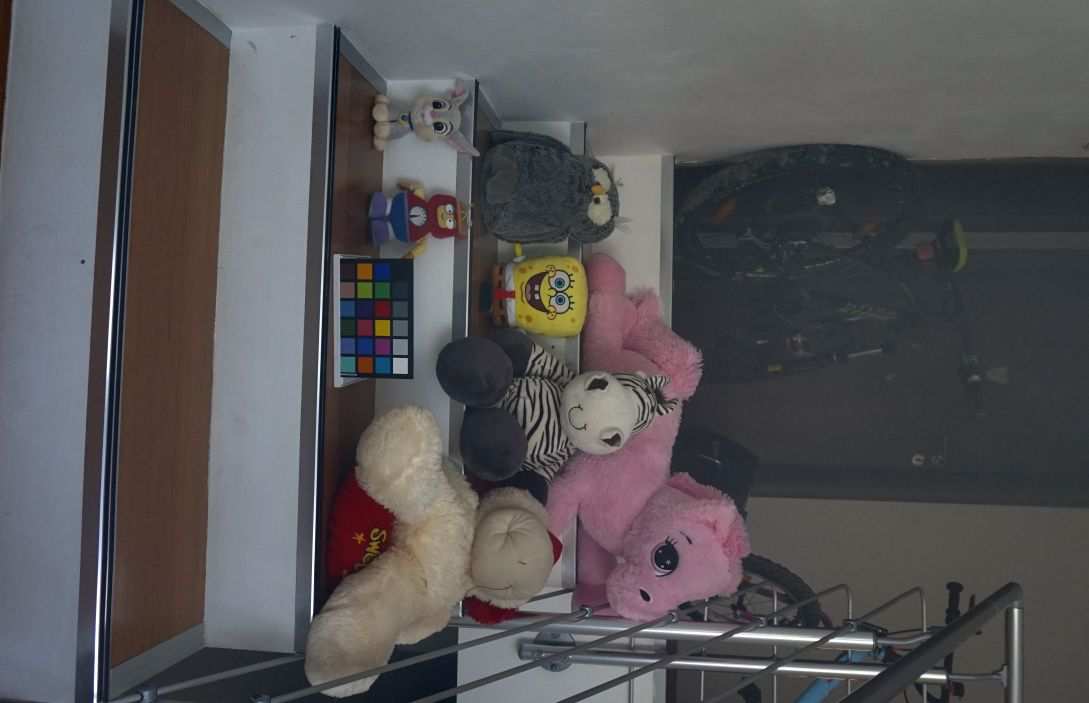} &
    \includegraphics[width=0.077 \linewidth,height=0.077 \linewidth]{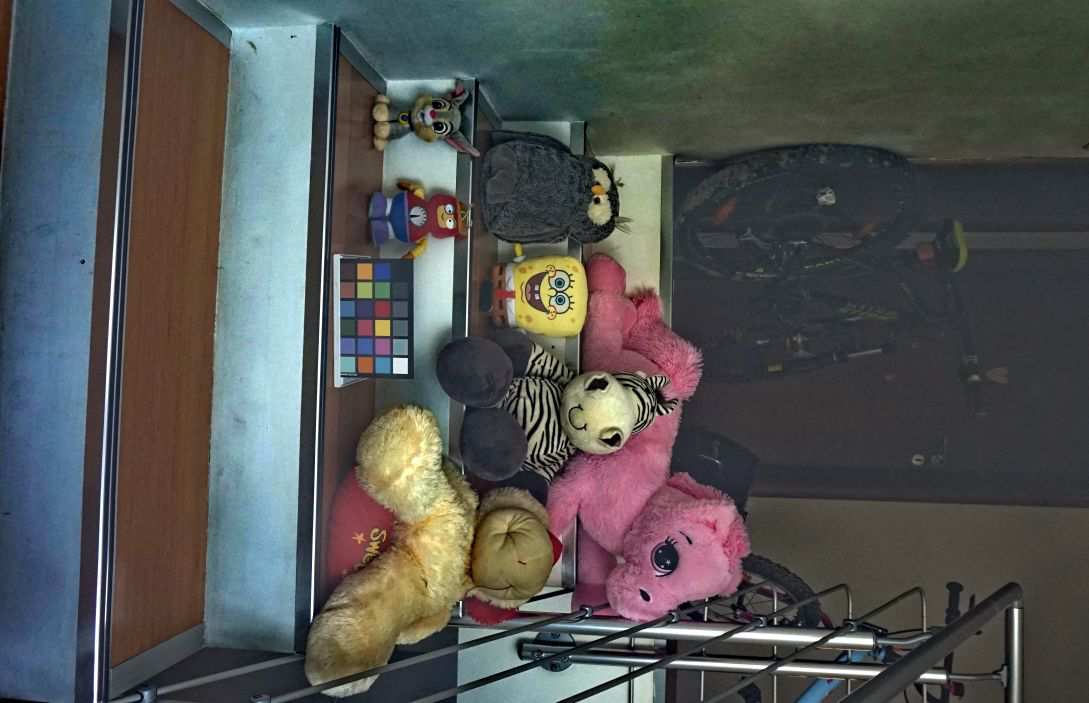} &  
    \includegraphics[width=0.077 \linewidth,height=0.077 \linewidth]{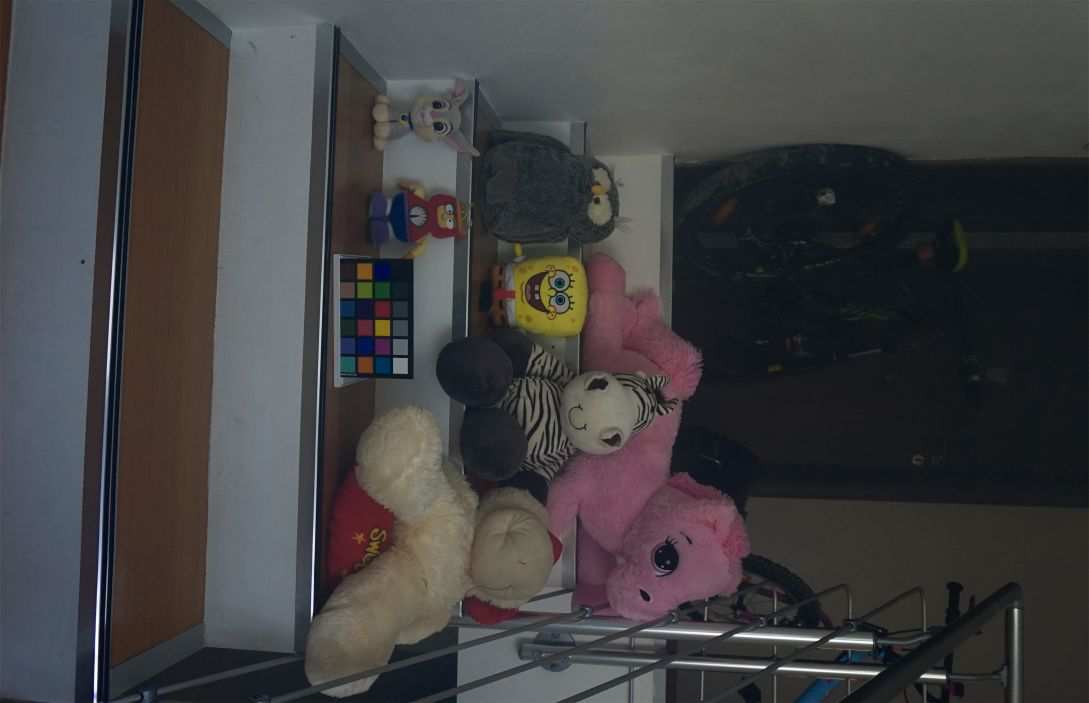} &
    \includegraphics[width=0.077 \linewidth,height=0.077 \linewidth]{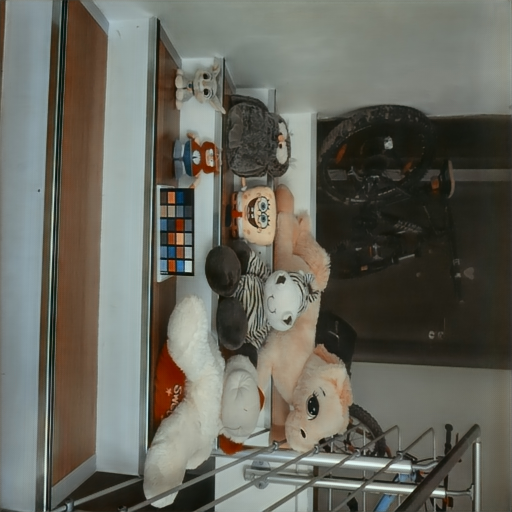} &  
    \includegraphics[width=0.077 \linewidth,height=0.077 \linewidth]{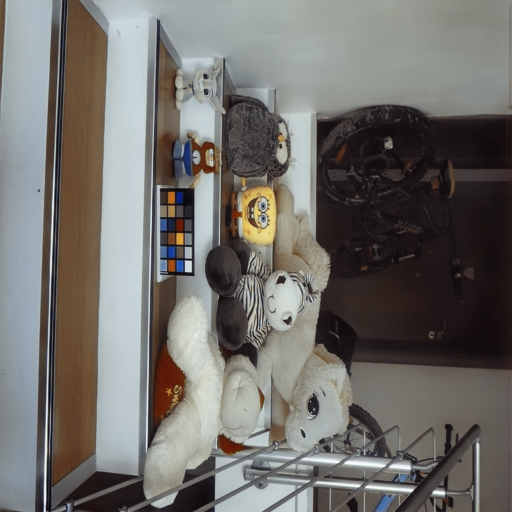} &
    \includegraphics[width=0.077 \linewidth,height=0.077 \linewidth]{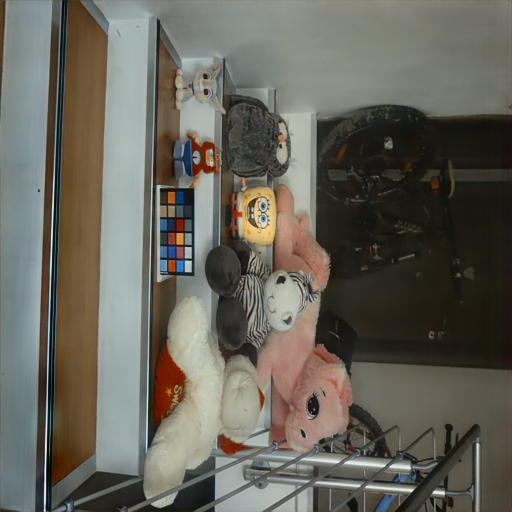} &
    \includegraphics[width=0.077 \linewidth,height=0.077 \linewidth]{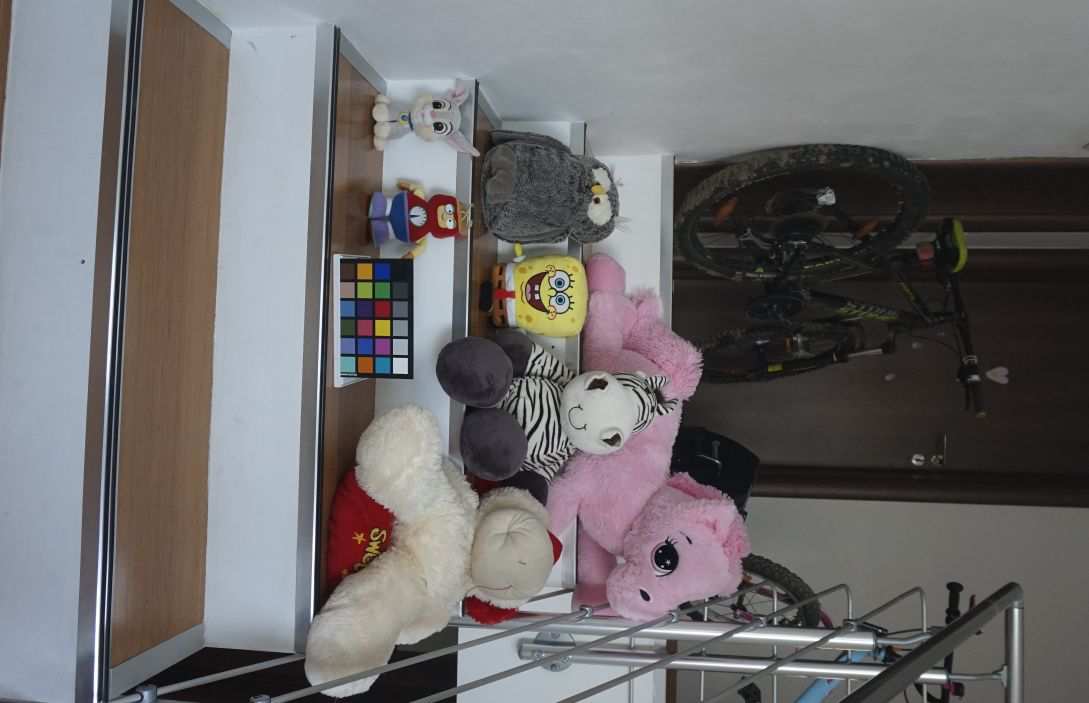} \\
    
\end{tabular}
}

\caption{Example input and ground-truth samples from the I-Haze dataset and the respective results of dehazing obtained using various methods. BPP results are presented with as well as withou the inclusion of our LGA module.}
\label{fig:dehazing}
\end{figure*}

\begin{table}[t]
\setlength{\tabcolsep}{03pt}	
\renewcommand{\arraystretch}{1}
\caption{Comparison of performance of different methods with our adaptation(BPPNet+LGA) in dehazing task on I-HAZE ~\cite{DBLP:journals/corr/abs-1804-05091} dataset.}
\centering
\resizebox{!}{!}{
\begin{tabular}{lcc}
\toprule
MODEL                   & SSIM ($\uparrow$)   & PSNR ($\uparrow$)  \\ \midrule
Input (hazy image)                   & 0.7302 & 13.80 \\
He \emph{et al.} \cite{he2010single}                & 0.7516 & 14.43 \\
Zhu \emph{et al.} \cite{zhu2015fast}                  & 0.6065 & 12.24 \\
Ren \emph{et al.} \cite{ren2016single}                 & 0.7545 & 15.22 \\
Berman \emph{et al.} \cite{berman2016non}                & 0.6537 & 14.12 \\
Li \emph{et al.} \cite{li2017all}                 & 0.7323 & 13.98 \\
BPPNet-reduced          & 0.8482 & 18.89 \\
BPPNet-reduced with LGA & \textbf{0.8663} & \textbf{20.17} \\ \bottomrule
\end{tabular}
}
\label{tab:dehazing}
\end{table}

\begin{figure}[t]
	\centering
	\includegraphics[width=.7\linewidth]{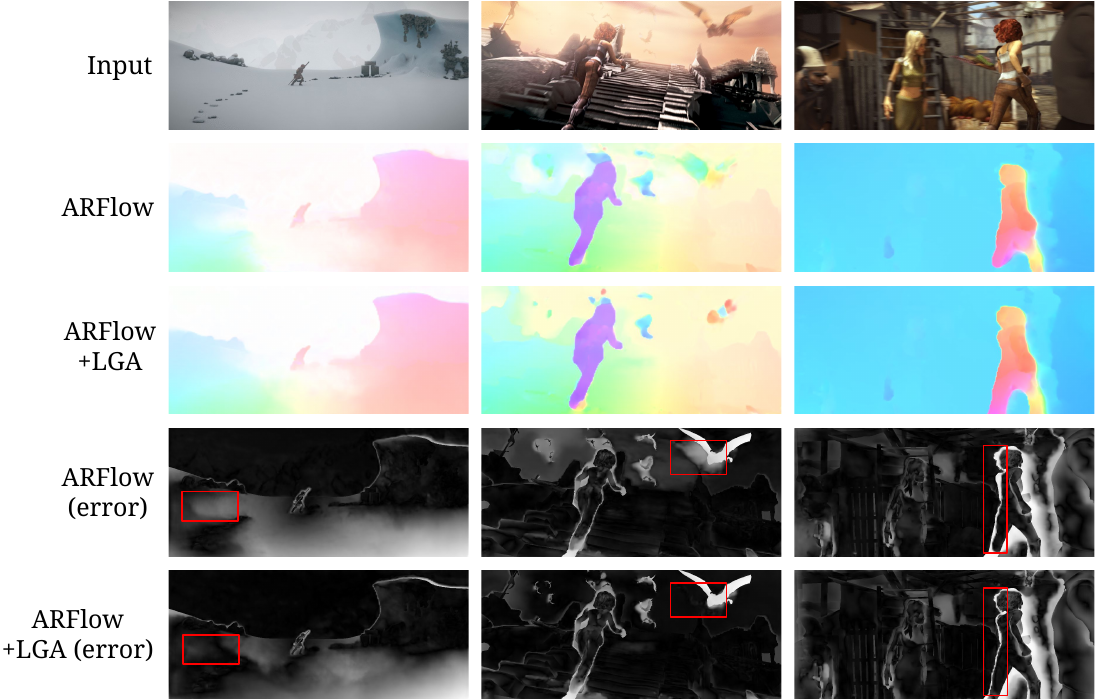}
	\caption{Results of unsupervised optical flow prediction on the MPI Sintel final pass benchmark \cite{butler2012naturalistic}. First row shows the input samples, then next two rows show estimated flow for ARFlow \cite{liu2020learning} and LGA with ARFlow. The last two rows visualize the incurred EPE-all errors for the methods on the final pass. Notable differences are highlighted in red box.}
	\label{fig:optical flow results}
\end{figure}

\begin{table}[t]
	\setlength{\tabcolsep}{03pt}	
	\renewcommand{\arraystretch}{1}
	\caption{Optical flow estimation comparison on the MPI Sintel benchmark \cite{butler2012naturalistic} on final and clean sets. The proposed LGA method achieved superior end-point error (EPE) scores on two-frame based unsupervised flow estimation task.}
	\centering
	\resizebox{!}{!}{
		\begin{tabular}{lcccccc}
			\toprule
			& \multicolumn{3}{c}{Final ($\downarrow$)}    & \multicolumn{3}{c}{Clean ($\downarrow$)}    \\ \cmidrule(r){2-4} \cmidrule(l){5-7}     
			MODEL        & \begin{tabular}[c]{@{}c@{}}EPE\\ all\end{tabular} & \begin{tabular}[c]{@{}c@{}}EPE\\ matched\end{tabular} & \begin{tabular}[c]{@{}c@{}}EPE\\ unmatched\end{tabular} & \begin{tabular}[c]{@{}c@{}}EPE\\ all\end{tabular} & \begin{tabular}[c]{@{}c@{}}EPE\\ matched\end{tabular} & \begin{tabular}[c]{@{}c@{}}EPE\\ unmatched\end{tabular} \\ \midrule
			UFlow \cite{jonschkowski2020matters}        & 6.498                                             & 3.078                                                 & 34.398                                                  & 5.205                                             & 2.036                                                 & 31.058                                                  \\
			FastFlowNet \cite{kong2021fastflownet}  & 6.080                                             & 2.942                                                 & 31.692                                                  & 4.886                                             & 1.789                                                 & 30.182                                                  \\
			MDFlow-fast \cite{kong2022mdflow}  & 5.994                                             & 2.770                                                 & 32.283                                                  & 4.733                                             & 1.673                                                 & 29.718                                                  \\
			UnsupSimFlow \cite{im2020unsupervised} & 6.916                                             & 3.017                                                 & 38.702                                                  & 5.926                                             & 2.159                                                 & 36.655                                                  \\
			ARFlow \cite{liu2020learning}      & 5.889                                             & 2.734                                                 & 31.602                                                  & 4.782                                             & 1.908                                                 & 28.261                                                  \\
			LGA           & \textbf{5.502}                                    & \textbf{2.604}                                        & \textbf{29.142}                                         & \textbf{4.109}                                    & \textbf{1.597}                                        & \textbf{24.626}                                         \\ \bottomrule
		\end{tabular}
	}
	\label{tab:optical flow final}
\end{table}

\subsection{Dehazing}
Most ehazing models are quite large in size and need large number of FLOPs. We compare several dehazing works with the BPPNet's reduced version with and without LGA (refer to the ablation study in Table \ref{tab:ablation4}). The quantitative results presented in Table \ref{tab:dehazing} show that the model with LGA performs the best in terms of SSIM. It also performs better than all the other methods in terms of PSNR. The significant reduction in computational cost of model with LGA was already established in Table \ref{tab:ablation4}. Therefore, it is evident that LGA supports computationally light and high-performance models for dehazing as well. The qualitative results of dehazing shown in Fig. \ref{fig:dehazing} also illustrate the superiority of image restoration achieved using LGA-incorporated reduced BPPNet model. 

\subsection{Optical flow estimation}
We incorporated LGA in ARFlow \cite{liu2020learning} which is an unsupervised optical flow estimation method. The results of two-frames based flow estimation are reported in Table \ref{tab:optical flow final} on MPI Sintel benchmark \cite{butler2012naturalistic} for final and clean sets using the standard end-point-error (EPE) metric. We see that ARFlow with LGA achieves superior performance in comparison to the other unsupervised methods on all six metrics. It surpasses ARFlow with an EPE-all of 0.387. Qualitative results are shown in Fig. \ref{fig:optical flow results}. As highlighted by the red bounding boxes, LGA-based method produces less error than ARFlow on the object edges, which shows LGA has a better understanding of moving and static background by identifying the non-local contexts.

\begin{table}
	\setlength{\tabcolsep}{03pt}	
	\renewcommand{\arraystretch}{1}
	\caption{Ablation study 1: Improvement in mIoU (\%) for different architectures after incorporating LGA.}
	\centering
	\resizebox{!}{!}{
		\begin{tabular}{lll}
			\toprule
			MODEL	       & Squeeze \cite{Squeeze1} & Shuffle \cite{zhang2017shufflenet} \\ \midrule
			Original model & 41.5    & 36.9    \\
			With LGA       & 44.6    & 44.5    \\ \bottomrule
		\end{tabular}
		\label{tab:ablation1}
	}
\end{table}

\subsection{Ablation Study}\label{sec:ablation}For the
study presented in Sec. \ref{sec:ablation_1}, \ref{sec:ablation_2} and \ref{sec:ablation_3}, we consider the transparent object segmentation problem, and for Sec. \ref{sec:ablation_4}, the problem of dehazing is considered. 

\subsubsection{Effect of incorporating LGA}\label{sec:ablation_1}
We consider the following architectures for encoder: Squeeze \cite{Squeeze1} and Shuffle \cite{zhang2017shufflenet}.
We evaluated the improvement in performance (in terms of mIoU \%) of both models when LGA module was incorporated in them. For Squeeze, the mIoU of the original model and after incorporating LGA module were 41.5\% and 44.6\%, respectively, indicating an improvement of more than 3\%. Shuffle benefited from including the LGA module significantly, with the improved mIoU being 45.5\% against 36.9\% of the original architecture.
We can conclude from the Table \ref{tab:ablation1} that the inclusion of LGA improves the performance of these networks. We also note that the problem of transparent object segmentation is quite challenging, with the best mIoU being $\sim 41\%$ and $\sim 45\%$ for Squeeze in the original form and after incorporating LGA, respectively.

\begin{table}
	\setlength{\tabcolsep}{03pt}	
	\renewcommand{\arraystretch}{1}
	\caption{Ablation study 2: Comparison of LGA (with simple \& group convolutions) vs. CCNet, with SqueezeNet encoder. 
		The original mIoU and accuracy without any LGA or CCNet module was 41.5 and 76.2 respectively.}
	\centering
	\resizebox{!}{!}{
		\begin{tabular}{lllll}
			\toprule
			\multicolumn{1}{c}{\multirow{2}{*}{MODEL}} & \multicolumn{2}{c}{Performance ($\uparrow$)} & \multicolumn{2}{c}{Extra resources ($\downarrow$)}                                  \\ \cmidrule(r){2-3} \cmidrule(l){4-5}
			\multicolumn{1}{c}{}                       & mIoU (\%)    & Accuracy (\%)    & Parameters ($\times 10^3$) & FLOPS ($\times 10^6$) \\ \midrule
			CCNet \cite{huang2020ccnet}                                        & 42.8         & 79.6             & 2686                                 & 5652                           \\
			LGA                                        & 45.8         & 81.8             & 132                                 & 140                           \\
			LGA small                & 44.6         & 79.6             & 17                               & 22                         \\ \bottomrule
		\end{tabular}
	}
	\label{tab:ablation2}
\end{table}

\subsubsection{Original SqueezeNet \cite{Squeeze1} with LGA or CCNet}\label{sec:ablation_2} We compared the performance of SqueezeNet architecture when CCNet or LGA is incorporated in it. The results are presented in Table \ref{tab:ablation2}. It is noted that as compared to original version, LGA indeed adds extra computation and storage demands but significantly lesser than CCNet. This has been discussed in Sec. \ref{sec:efficiency} as well. Moreover, it is noteworthy that the performance of SqueezeNet with LGA is better than with CCNet in terms of both mIoU and accuracy. 

\subsubsection{Effect of divergence loss and group convolution}\label{sec:ablation_3} We performed a qualitative evaluation of different models with LGA module via training them with and without the proposed divergence loss. The results presented in Table \ref{tab:ablation3} show that all the models witness a consistent improvement when the divergence loss is introduced. In addition, we note that the configuration for the best performance (LGA using simple convolution and with divergence loss) requires only marginally extra number of parameters and computations over the original SqueezeNet without LGA. Using group-transpose convolution strategy instead of simple convolution in message propagation further sheds down the storage and computation needs, with only a minor drop in the performance. The uses of Group-convolution reduces the computational cost term $NC^2$ as mentioned in Sec. \ref{sec:efficiency}. Similar observation is made also for Shuffle-Net.

\begin{table}
	\setlength{\tabcolsep}{03pt}	
	\renewcommand{\arraystretch}{1}
	\caption{Ablation study 3: Comparison of mIOU (\%) of LGA using simple convolution (SC) or group convolution (GC), and with or without divergence loss. We also include the number of parameters and FLOPs for the first three rows to illustrate that the computational needs of incorporating LGA are marginal, and even lesser if we employ group-transposed convolution.}
	\centering
	\resizebox{!}{!}{
		\begin{tabular}{lcccc}
			\toprule
			MODEL                    & Variation            & mIoU (\%)           & \begin{tabular}[c]{@{}c@{}}Parameters\\ ($\times 10^6$)\end{tabular} & \begin{tabular}[c]{@{}c@{}}FLOPS\\ ($\times 10^9$)\end{tabular} \\ \midrule
			\multirow{5}{*}{Squeeze} & No LGA               & 41.5                & 1.018                               & 13.140                          \\
			& LGA with SC and $L_{div}$ & {\underline{\textbf{45.8}}} & 1.150                               & 13.280                          \\
			& LGA with GC and $L_{div}$ & \textbf{44.6}       & 1.035                               & 13.162                          \\
			& LGA with SC, no $L_{div}$ & 43.6                &                                     &                                \\
			& LGA with GC, no $L_{div}$ & 42.3                &                                     &                                \\ \cmidrule{2-5} 
			\multirow{5}{*}{Shuffle} & No LGA               & 36.9                & 0.395                               & 3.42                           \\
			& LGA with SC and $L_{div}$ & {\underline{\textbf{44.6}}} & 0.506                               & 3.69                           \\
			& LGA with GC and $L_{div}$ & \textbf{44.5}                & 0.412                               & 3.5                            \\
			& LGA with SC, no $L_{div}$ & 43.0                &                                     &                                \\
			& LGA with GC, no $L_{div}$ & 41.6                &                                     &                                \\ \bottomrule
		\end{tabular}
	}
	\label{tab:ablation3}
\end{table}

\subsubsection{Ablation on BPPNet for dehazing problem}\label{sec:ablation_4} We compared the performance and computational load of the original BPPNet \cite{singh2020single}, the reduced BPPNet (as explained in Sec. \ref{subsec:training process}) without LGA, and reduced BPPNet with LGA. The results are reported in Table \ref{tab:ablation4}. It is seen that the reduced BPPNet without LGA has less than 10\% number of parameters of the original and approximately 20\% number of FLOPs. Indeed, this resulted in poorer performance in terms of both structural similarity index (SSIM) and peak-signal-to-noise-ratio (PSNR). However, the inclusion of LGA improves both the SSIM and the PSNR, with almost no extra computation load.

\begin{table}
\setlength{\tabcolsep}{03pt}	
\renewcommand{\arraystretch}{1}
\caption{Ablation study 4: Ablation on BPPNet~\cite{singh2020single} for dehazing of I-HAZE ~\cite{DBLP:journals/corr/abs-1804-05091} dataset.}
\centering
\resizebox{!}{!}{
\begin{tabular}{lccc}
\toprule
& Original & Reduced & Reduced with LGA \\ \midrule
SSIM                                & 0.8994   & 0.8482  & 0.8663           \\
PSNR                                & 22.56    & 18.89   & 20.17            \\
Parameters ($\times 10^6$) & 8.851    & 0.685   & 0.687            \\
FLOPS ($\times 10^9$)      & 348.49   & 87.44   & 87.78            \\ \bottomrule
\end{tabular}
}
\label{tab:ablation4}
\end{table}

\subsubsection{Ablation for number of LGA layers}\label{sec:ablation_5}
The ablation is given is given in Table \ref{tab:layer_ab1}, mIoU increases continuously till a certain number of layers as the region of spatial context increases, and then falls marginally due to over-smoothing effect. We have taken a fixed kernel size 9. As from kernel size, we understand a number of edge connections from a single node. This is not a hyperparameter in LGA, but a design choice based on neighborhood. Smaller number of kernels inhibit the propagation of information to the nearest neighbor. If so, LGA needs multiple layers to reach the neighbor excluded in connectivity. Larger number of kernels implies redundancy.

\begin{table}
\setlength{\tabcolsep}{03pt}	
\renewcommand{\arraystretch}{1}
\caption{Ablation for no. of LGA layers on Trans10Kv2(Squeeze).}
\centering
\resizebox{!}{!}{
\begin{tabular}{lccccc}
\toprule
No. of LGA layers & 0    & 1    & 2    & 4    & 8    \\ \midrule
mIoU (\%)         & 36.9 & 43.1 & 43.9 & 44.5 & 41.1 \\ \bottomrule
\end{tabular}
\label{tab:layer_ab1}
}
\end{table}

\begin{table*}[t]
	\setlength{\tabcolsep}{03pt}	
	\renewcommand{\arraystretch}{1}
	\caption{Ablation study on performance and efficiency comparison of LGA vs. CCNet \cite{huang2020ccnet}, both with SqueezeNet encoder, on Trans10Kv2 dataset \cite{translab2}. Here, LGA means with simple convolutions and LGA small means with group convolutions.
		The original mIoU and accuracy of SqueezeNet \cite{Squeeze1} without any LGA or CCNet module were 41.5 and 76.2 respectively.}
	\centering
	\resizebox{!}{!}{
		\begin{tabular}{lccccccccc}
			\toprule
			& \multicolumn{2}{c}{Performance ($\uparrow$)}                   & \multicolumn{3}{c}{Extra Parameters ($10^3$) ($\downarrow$)}                                                                                                                                         & \multicolumn{4}{c}{Extra FLOPS ($10^6$) ($\downarrow$)}                                                                                                                                                                                                        \\ \cmidrule(r){2-3} \cmidrule(lr){4-6} \cmidrule(l){7-10}  
			\multirow{2}{*}{MODEL} & \multirow{2}{*}{mIoU} & \multirow{2}{*}{Accuracy} & \multirow{2}{*}{\begin{tabular}[c]{@{}c@{}}Conv. channel\\ resizing\end{tabular}} & \multirow{2}{*}{\begin{tabular}[c]{@{}c@{}}Attention\\ module\end{tabular}} & \multirow{2}{*}{Total} & \multirow{2}{*}{\begin{tabular}[c]{@{}c@{}}Conv. channel\\ resizing\end{tabular}} & \multicolumn{2}{c}{Attention module}                                                                                                 & \multirow{2}{*}{Total} \\ \cmidrule(lr){8-9}
			&                       &                           &                                                                                   &                                                                             &                        &                                                                                   & \begin{tabular}[c]{@{}c@{}}Information\\ propagation\end{tabular} & \begin{tabular}[c]{@{}c@{}}Other Conv.\\ operations\end{tabular} &                        \\ \hline
			CCNet \cite{huang2020ccnet}                  & 42.8                  & 79.6                      & 2359                                                                              & 327                                                                         & 2686                   & 4832                                                                              & 150                                                               & 670                                                              & 5652                   \\
			LGA                    & 45.8                  & 81.8                      & 66                                                                                & 67                                                                          & 132                    & 67                                                                                & 5                                                                 & 68                                                               & 140                    \\
			LGA small              & 44.6                  & 79.6                      & 8                                                                                 & 9                                                                           & 17                     & 8                                                                                 & 5                                                                 & 9                                                                & 22                     \\ \bottomrule

		\end{tabular}
	}
	\label{tab:ablation2 extended}
\end{table*}

\subsubsection{Performance and efficiency comparison of LGA vs. CCNet}
\label{sec:lga_vs_ccnet}
Table \ref{tab:ablation2 extended} presents comparison of CCNet \cite{huang2020ccnet} with the proposed LGA module on Trans10Kv2 dataset \cite{translab2}. Performance of LGA is better than CCNet in terms of both mIoU and accuracy. Even after using group convolutions in LGA, the performance is better than CCNet with a significant reduction in space and time complexity. Among all three, LGA with group convolution needs orders of magnitude less extra parameters and computation. For both parameters and computation, convolution channel resizing is the costliest operation, much more than the attention module. By using a graph based approach, LGA saves a significant amount of resources.

\section{Conclusion}
This paper presents a novel concept of latent graph attention (LGA). The computational efficiency of LGA and its contribution to performance improvement and versatility is demonstrated through multiple studies and three example applications within the image-to-image translation domain. It is shown to be an effective mechanism to learn global context and contribute this information to the parent architecture in which it is included as a module. We hope that LGA will find use in many challenging applications and across a wide variety of architectures where it is not practical or straightforward to incorporate attention or global context mechanisms. In particular, we expect that LGA will be a powerful tool to upgrade the performance of edge devices with very limited memory and computation resources. We will release the source code, models, and LGA libraries on acceptance.

\appendix



\section{Complexity analysis}
\label{sec:complexity analysis}

\subsection{Time complexity}
\label{subsec:time complexity}
This section provides the derivation of time complexity of LGA network in comparison to the best light weight criss-cross attention network (CCNet \cite{huang2020ccnet}). The time complexity of the attention networks can be defined as

\begin{equation} \label{eq:time complexity 1}
	{
		\begin{split}
			\mathcal{O}\big(Conv Transformation\big) + \mathcal{O}\big(LGA/CCNet \; attention\big)
		\end{split}
	}
\end{equation}

$\mathcal{O}\big(Conv Transformation\big)$ is the time required to do channel transformation using convolution layer. For example, changing channel size to reduce overall complexity. For LGA, it can be written as

\begin{equation} \label{eq:time complexity 2}
	{
		\begin{split}
			\mathcal{O}\big(Conv Transformation\big) & = a_1 \times N \times C_{in} \times C_{out} \times k \times k \\
			& + b_1 \times N\times C_{out}\times C_{out}\times k \times k
		\end{split}
	}
\end{equation}

Here, $k=1$ is the kernel size, $a_1$ and $b_1$ are number of times such operations are applied. $N = n\times n$ is the feature shape. Equation \eqref{eq:time complexity 2} can be simplified as

\begin{equation} \label{eq:time complexity 3}
	{
		\begin{split}
			\mathcal{O}\big(Conv Transformation\big) & = a_1 \times N \times C_{in} \times C_{out} \\
			& + b_1 \times N\times C_{out}\times C_{out}
		\end{split}
	}
\end{equation}

Here, $C_{in} > C_{out}$, $a_1 = 1$ and $b_1 > a_1$. Using these conditions, we can approximate that $a_1\times C_{in} \approx b_1\times C_{out}$.

\begin{equation} \label{eq:time complexity 4}
	{
		\begin{split}
			\therefore \mathcal{O}\big(Conv Transformation\big) = 2\times b\times N\times C_{out}\times C_{out}
		\end{split}
	}
\end{equation}

\begin{equation} \label{eq:time complexity 5}
	{
		\begin{split}
			\mathcal{O}\big(LGA \; attention\big) = c_1 \times N \times C_{out}
		\end{split}
	}
\end{equation}

\begin{equation} \label{eq:time complexity 6}
	{
		\begin{split}
			\mathcal{O}\big(LGA\big) & = b_1\times N\times C_{out}\times C_{out} \\
			& + c_1 \times N \times C_{out}
		\end{split}
	}
\end{equation}

as $b_1$ and $c_1$ are small, we can rewrite Eq. \eqref{eq:time complexity 6} as

\begin{equation} \label{eq:time complexity 7}
	{
		\begin{split}
			\mathcal{O}\big(LGA\big) & = N\times C_{out}\times C_{out} \\
			& + N \times C_{out}
		\end{split}
	}
\end{equation}

Now, for CCNet we have

\begin{equation} \label{eq:time complexity 8}
	{
		\begin{split}
			\mathcal{O}\big(Conv Transformation\big) & = a_2 \times N \times C_{in} \times C_{out} \times k \times k \\
			& + b_2 \times N\times C_{in}\times C_{in}\times k \times k
		\end{split}
	}
\end{equation}

\begin{equation} \label{eq:time complexity 9}
	{
		\begin{split}
			\mathcal{O}\big(CCNet \; attention\big) = c_2 \times N^{3/2} \times C_{out}
		\end{split}
	}
\end{equation}

\begin{equation} \label{eq:time complexity 10}
	{
		\begin{split}
			\mathcal{O}\big(CCNet\big) & = N\times C_{in}\times C_{out} \\
			& + N\times C_{in}\times C_{in} \\
			& + N^{3/2} \times C_{out}
		\end{split}
	}
\end{equation}

considering $a_2$, $b_2$, $c_2$, $k$ as small. As $C_{in} > C_{out}$, we can rewrite Eq. \eqref{eq:time complexity 10} as

\begin{equation} \label{eq:time complexity 11}
	{
		\begin{split}
			\mathcal{O}\big(CCNet\big) & = N\times C_{in}\times C_{in} \\
			& + N^{3/2} \times C_{out}
		\end{split}
	}
\end{equation}

By comparing time complexities of CCNet and LGA in Eq. \eqref{eq:time complexity 7} and Eq. \eqref{eq:time complexity 11}, we learn that CCNet has $C_{in} > C_{out}$ and $N^{3/2}$ as compared to $N$ in LGA. Hence, it has higher FLOPs than LGA.

We found that it is possible to reduce the complexity of CCNet by adding a reducing Conv. layer. This makes the CCNet time complexity

\begin{equation} \label{eq:time complexity 12}
	{
		\begin{split}
			\mathcal{O}\big(CCNet\big) & = N\times C_{out}\times C_{out} \\
			& + N^{3/2} \times C_{out}
		\end{split}
	}
\end{equation}

Convolutional transformation term, $N\times C_{out}\times C_{out}$, can be reduced further by adding group convolutions. It modifies the time complexities as

\begin{equation} \label{eq:time complexity 13}
	{
		\begin{split}
			\mathcal{O}\big(CCNet\big) & = \frac{N\times C_{out}\times C_{out}}{G}
			+ N^{3/2} \times C_{out} \\
			\mathcal{O}\big(LGA\big) & = \frac{N\times C_{out}\times C_{out}}{G}
			+ N \times C_{out}
		\end{split}
	}
\end{equation}

where $G$ denotes the number of groups. If $G$ is large, $\frac{N\times C_{out}\times C_{out}}{G}$ becomes small. Then the bottleneck remains the attention term which cannot be reduced.

\begin{equation} \label{eq:time complexity 14}
	{
		\begin{split}
			\mathcal{O}\big(CCNet\big) & = N^{3/2} \times C_{out} \\
			\mathcal{O}\big(LGA\big) & =   N \times C_{out}
		\end{split}
	}
\end{equation}

Hence, in general the time complexity of attention of CCNet is $\sqrt{N}$ times higher than LGA attention.

\subsection{Space complexity}
\label{subsec:space complexity}
The space complexity of the attention based networks can be defined as Eq. \eqref{eq:time complexity 1}. For LGA, we have

\begin{equation} \label{eq:space complexity 15}
	{
		\begin{split}
			\mathcal{O}\big(Conv Transformation\big) & = a_1 \times N \times C_{in} \\
			& + b_1 \times N\times C_{out}
		\end{split}
	}
\end{equation}

As $b_1 > a_1$ and $C_{in} > C_{out}$, $a_1 = 1$ implies that $a_1\times C_{in} \approx b_1 \times C_{out}$. Using this, Eq. \eqref{eq:space complexity 15} can be rewritten as

\begin{equation} \label{eq:space complexity 16}
	{
		\begin{split}
			\mathcal{O}\big(Conv Transformation\big) & = a_1 \times N \times C_{in} \\
			Or, \;\; \mathcal{O}\big(Conv Transformation\big) & = \times N\times C_{in}
		\end{split}
	}
\end{equation}

As the adjacency matrix is sparse, so the $\mathcal{O}\big(LGA \; attention\big) = N$, and LGA space complexity becomes

\begin{equation} \label{eq:space complexity 17}
	{
		\begin{split}
			\mathcal{O}\big(LGA\big) = \times N\times C_{in} + N
		\end{split}
	}
\end{equation}

Now, for CCNet we have

\begin{equation} \label{eq:space complexity 18}
	{
		\begin{split}
			\mathcal{O}\big(Conv Transformation\big) & = a_1 \times N \times C_{in} \\
			& + b_1 \times N\times C_{out} \\
			& \approx a_1 \times N \times C_{in} \\
			& = N \times C_{in}
		\end{split}
	}
\end{equation}

Please note $N\times C_{in}$ cannot be reduced further because it is the input feature size which is dependent on the used encoder and not on the proposed LGA module. Also, if $a_1\times N \times C_{in} \neq b1\times N\times C_{out}$ and $b_1\times N \times C_{out} < a_1 \times N \times C_{in} $ with $(a_1 = 1)$ then as $N\times C_{in}$ is from encoder, $N\times C_{in}$ is the bottleneck and it cannot be reduced. Hence, $N\times C_{in}$ is the lower bound on $\mathcal{O}\big(Conv Transformation\big)$, hence, we have

\begin{equation} \label{eq:time complexity 19}
	{
		\begin{split}
			\mathcal{O}\big(CCNet \; attention\big) & = N^{3/2} \\
			Or, \;\; \mathcal{O}\big(CCNet\big) & = N\times C_{in} + N^{3/2}
		\end{split}
	}
\end{equation}

Comparing the space complexity of LGA with CCNet, we have

\begin{equation} \label{eq:time complexity 20}
	{
		\begin{split}
			\mathcal{O}\big(LGA\big) & = N\times C_{in} + N \\
			\mathcal{O}\big(CCNet\big) & = N\times C_{in} + N^{3/2}
		\end{split}
	}
\end{equation}

$N\times C_{in}$ term is same for LGA and CCNet. If $N$ is large then $N\times C_{in} + N \approx N$ and the space complexities become

\begin{equation} \label{eq:time complexity 20}
	{
		\begin{split}
			\mathcal{O}\big(LGA\big) & = N \\
			\mathcal{O}\big(CCNet\big) & = N^{3/2}
		\end{split}
	}
\end{equation}


\section{LGA contrastive loss analysis}
\label{sec:LGA contrastive loss analysis}

This novel loss term helps in the learning of our LGA module. Mathematically, LGA contrastive loss can be stated as

\begin{equation} \label{eq:divloss}
	{
		\begin{split}
			\mathcal{L}_{\rm LGA} = E_{P_i,P_j} \Bigg( \mathcal{C}_{ij}\log \left( \frac{V^{2}_{ij}}{U^{}_{ij}}  + 1\right)
			+ \bar{\mathcal{C}}_{ij}\log \left(
			\frac{U^{}_{ij}}{V^{2}_{ij}}  + 1\right)
			\Bigg)
		\end{split}
	}
\end{equation}

$V_{ij} = D(F^{\rm out}_i,F^{\rm out}_j)$ and $U_{ij} = D(F^{\rm in}_i,F^{\rm in}_j)$, $F^{\rm in}$ and $F^{\rm out}$ denote the input and output feature maps for LGA module, and $D(\cdot)$ is the divergence function such as KL-divergence or mean square error. The two variables $\mathcal{C}$ and $\bar{\mathcal C}$ are boolean, and their values depend on the similarity between the neighboring nodes in the graph. For example, for nodes $i$ and $j$, we look at the GT labels of patches containing them. Further, we aggregate the labels to assign a single label to the node. For example, one aggregate measure could be to assign the majority class label to the node. Let $Agg(\cdot)$ denote the aggregate function and $i$ and $j$ denote two nodes from the graph, then $\mathcal{C} = 1$ if $Agg(i) = Agg(j)$, and 0 otherwise.

Next we present the interpretation and consequence of using this loss function. We know that having only $V_{ij}$ will not be sufficient to enforce learning in LGA. To fulfill this, $U_{ij}$ is incorporated in the proposed contrastive loss.

We note that $U_{ij}$ is determined by the input to LGA module and does not update as LGA learns. LGA essentially learns through tweaking the value of $V_{ij}$, which is determined by the output of LGA. We further note that larger diverse value $D_{ij}$ indicates larger difference in the distributions of the nodes $i$ and $j$. With these points, we now assess how the loss function behaves and influences the learning of LGA in different situations. For the case when two nodes have a similar distribution $V_{ij}$ needs to be minimized, implying that the two nodes have similar output distributions as well. For cases where the input distributions of the two nodes are very different, $V_{ij}$ is learnt to be maximized thereby setting the two output distributions apart.

\paragraph{Case: similar distribution} Let us first consider the case of two nodes with similar distribution, such that only the first term (i.e. term with ${\mathcal{C}}_{ij}$) contributes to $\mathcal{L}_{\rm LGA}$. In this condition, we want divergence between nodes to be lower. If $U_{ij}$ is large and $V_{ij}$ is small, then the net value of $\mathcal{L}_{\rm LGA}$ is low. This indicates that LGA is able to identify that the nodes have similar distribution, even if their input features do not indicate so. 
On the other hand if $U_{ij}$ is small and $V_{ij}$ is large, then $\mathcal{L}_{\rm LGA}$ is large. This drives LGA layers to update themselves for reducing $V_{ij}$. This situation actually indicates that the network preceding LGA is able to identify that the nodes have similar distributions, but LGA should learn to propagate this property ahead.
If both $U_{ij}$ and $V_{ij}$ are high, the second power of $V_{ij}$ in the numerator helps LGA focus on learning to reduce $V_{ij}$, albeit at a slower pace than the previous situation (i.e. small $U_{ij}$). This is because in this situation both the LGA and the network preceding it have the scope of improvement, and the slow pace of update of LGA is an indirect nudge to the preceding or succeeding networks to improve their performance. In the ideal situation, when both $U_{ij}$ and $V_{ij}$ are low, $\mathcal{L}_{\rm LGA}$ remains low and LGA is effective in propagating the correctly learnt attributes of the preceding network.

\paragraph{Case: different distribution} Now, let us consider the second case when the nodes $i$ and $j$ do not have the same distribution. $\mathcal{L}_{\rm LGA}$ then corresponds to only the second term (with $\bar{\mathcal{C}}_{ij}$). In this condition, we want divergence between nodes to be high. If $U_{ij}$ is large but $V_{ij}$ is small, then the value of $\mathcal{L}_{\rm LGA}$ is large. Then LGA updates itself to increase the value of $V_{ij}$. This also indicates that the preceding network is effective in separating the $F^{\rm in}$ of the two nodes and the LGA needs to learn to propagate this attribute of the preceding network efficiently. On the other hand if $U_{ij}$ is small but $V_{ij}$ is large, the loss function has a small value due to the second power of $V_{ij}$ in the denominator, which indicates that LGA is able to learn that the nodes have different distributions despite the poor performance of the preceding network. 
In the situation that both $U_{ij}$ and $V_{ij}$ are low, $\mathcal{L}_{\rm LGA}$ is high due to the second power of $V_{ij}$ in the denominator. Then, LGA learns to increase the value of $V_{ij}$ in a relatively slow and linear manner. This is so because this situation indicates that there is a scope of improvement for both LGA and the network preceding it. This slow update of LGA can provide a nudge to the preceding or succeeding networks to improve their learning as well.

\paragraph{Case: ideal distribution} Lastly, in the ideal situation where both $U_{ij}$ and $V_{ij}$ are large, the loss function is low owing to the second order of $V_{ij}$ in the denominator. This indicates that LGA has learned to propagate the good performance of the preceding network in separating the two nodes in an effective manner.


\section{Architectural details}
\label{sec:architectural details}

Table \ref{tab:arch seg squeeze}, \ref{tab:arch seg shuffle}, \ref{tab:arch dehazing}, \ref{tab:arch optical flow} present the architectural details of LGA module when applied to the three sets of problems on SqueezeNet \cite{Squeeze1}, ShuffleNet \cite{zhang2017shufflenet}, BPPNet \cite{singh2020single} and ARFlow \cite{liu2020learning}. The tables describe operations at each step, associated kernels, inputs and outputs.

\begin{table}[t]
	\setlength{\tabcolsep}{03pt}	
	\renewcommand{\arraystretch}{1}
	\caption{Architecture details of incorporation of LGA into SqueezeNet \cite{Squeeze1} for segmentation. Starting from the input $F^{\rm in}$, till the LGA output after concatenation.}
	\centering
	\resizebox{!}{!}{
		\begin{tabular}{lccccc}
			\toprule
			& Size of tensor & Input size                                                        & \begin{tabular}[c]{@{}c@{}}Conv. parameter (input\\ \& output channel)\end{tabular} & Output size & Groups \\ \midrule
			$F^{\rm in}$           & 32x32x512      &                                                                   &                                                                                      &             &        \\
			(1x1) 2d conv   &                & 32x32x512                                                         & 512,128                                                                              & 32x32x128   & 8      \\
			9 (1x1) 2d conv &                & 32x32x128                                                         & 128,1                                                                                &             & 1      \\
			Adj. matrix     & 1024x1024      &                                                                   &                                                                                      &             &        \\
			(1x1) 1d conv   &                & 1024x128                                                          & 128,128                                                                              & 1024x128    & 8      \\
			(1x1) 1d conv   &                & 1024x128                                                          & 128,128                                                                              & 1024x128    & 8      \\
			(1x1) 1d conv   &                & 1024x128                                                          & 128,128                                                                              & 1024x128    & 8      \\
			(1x1) 1d conv   &                & 1024x128                                                          & 128,128                                                                              & 1024x128    & 8      \\
			Concat          &                & \begin{tabular}[c]{@{}c@{}}32x32x512 \&\\ 32x32x128\end{tabular} &                                                                                      & 32x32x640   &        \\
			Output          & 32x32x640      &                                                                   &                                                                                      &             &        \\ \bottomrule
		\end{tabular}
	}
	\label{tab:arch seg squeeze}
\end{table}

\begin{table}[t]
	\setlength{\tabcolsep}{03pt}	
	\renewcommand{\arraystretch}{1}
	\caption{Architecture details of incorporation of LGA into ShuffleNet \cite{zhang2017shufflenet} for segmentation. Starting from the input $F^{\rm in}$, till the LGA output after concatenation.}
	\centering
	\resizebox{!}{!}{
		\begin{tabular}{lccccc}
			\toprule
			& Size of tensor & Input size                                                        & \begin{tabular}[c]{@{}c@{}}Conv. parameter (input\\ \& output channel)\end{tabular} & Output size & Groups \\ \midrule
			$F^{\rm in}$           & 32x32x232      &                                                                   &                                                                                      &             &        \\
			(1x1) 2d conv   &                & 32x32x232                                                         & 232,136                                                                              & 32x32x136   & 8      \\
			9 (1x1) 2d conv &                & 32x32x128                                                         & 128,1                                                                                &             & 1      \\
			Adj. matrix     & 1024x1024      &                                                                   &                                                                                      &             &        \\
			(1x1) 1d conv   &                & 1024x136                                                          & 136,136                                                                              & 1024x136    & 8      \\
			(1x1) 1d conv   &                & 1024x136                                                          & 136,136                                                                              & 1024x136    & 8      \\
			(1x1) 1d conv   &                & 1024x136                                                          & 136,136                                                                              & 1024x136    & 8      \\
			(1x1) 1d conv   &                & 1024x136                                                          & 136,136                                                                              & 1024x136    & 8      \\
			Concat          &                & \begin{tabular}[c]{@{}c@{}}32x32x232 \&\\ 32x32x136\end{tabular} &                                                                                      & 32x32x368   &        \\
			Output          & 32x32x368      &                                                                   &                                                                                      &             &        \\ \bottomrule
		\end{tabular}
	}
	\label{tab:arch seg shuffle}
\end{table}

\begin{table}[t]
	\setlength{\tabcolsep}{03pt}	
	\renewcommand{\arraystretch}{1}
	\caption{Architecture details of incorporation of LGA into BPPNet \cite{singh2020single} for dehazing. Starting from the input $F^{\rm in}$, till the LGA output after concatenation.}
	\centering
	\resizebox{!}{!}{
		\begin{tabular}{lccccc}
			\toprule
			& Size of tensor & Input size                & \begin{tabular}[c]{@{}c@{}}Conv. parameter (input\\ \& output channel)\end{tabular} & Output size & Groups \\ \midrule
			$F^{\rm in}$           & 512x512x16     &                           &                                                                                      &             &        \\
			resize          &                & 512x512x16                &                                                                                      & 32x32x16    &        \\
			9 (1x1) 2d conv &                & 32x32x16                  & 16,1                                                                                 &             & 1      \\
			Adj. matrix     & 1024x1024      &                           &                                                                                      &             &        \\
			(1x1) 1d conv   &                & 16x16                     & 16,16                                                                                & 16x16       & 1      \\
			(1x1) 1d conv   &                & 16x16                     & 16,16                                                                                & 16x16       & 1      \\
			(1x1) 1d conv   &                & 16x16                     & 16,16                                                                                & 16x16       & 1      \\
			(1x1) 1d conv   &                & 16x16                     & 16,16                                                                                & 16x16       & 1      \\
			resize          &                & 32x32x16                  &                                                                                      & 512x512x16  &        \\
			Concat          &                & \begin{tabular}[c]{@{}c@{}}512x512x16 \&\\  512x512x16\end{tabular} &                                                                                      & 512x512x32  &        \\
			Output          & 512x512x32     &                           &                                                                                      &             &        \\ \bottomrule
		\end{tabular}
	}
	\label{tab:arch dehazing}
\end{table}

\begin{table}[t]
	\setlength{\tabcolsep}{03pt}	
	\renewcommand{\arraystretch}{1}
	\caption{Architecture details of incorporation of LGA into ARFlow \cite{liu2020learning} for optical flow estimation. Starting from the input $F^{\rm in}$, till the LGA output after concatenation.}
	\centering
	\resizebox{!}{!}{
		\begin{tabular}{lccccc}
			\toprule
			& Size of tensor & Input size                                                      & \begin{tabular}[c]{@{}c@{}}Conv. parameter (input\\ \& output channel)\end{tabular} & Output size & Groups \\ \midrule
			$F^{\rm in}$           & 24x28x96       &                                                                 &                                                                                      &             &        \\
			resize          &                & 24x28x96                                                        &                                                                                      & 32x32x96    &        \\
			9 (1x1) 2d conv &                & 32x32x96                                                        & 96,1                                                                                 &             & 1      \\
			Adj. matrix     & 1024x1024      &                                                                 &                                                                                      &             &        \\
			(1x1) 1d conv   &                & 96x96                                                           & 96,96                                                                                & 96x96       & 8      \\
			(1x1) 1d conv   &                & 96x96                                                           & 96,96                                                                                & 96x96       & 8      \\
			(1x1) 1d conv   &                & 96x96                                                           & 96,96                                                                                & 96x96       & 8      \\
			(1x1) 1d conv   &                & 96x96                                                           & 96,96                                                                                & 96x96       & 8      \\
			resize          &                & 32x32x96                                                        &                                                                                      & 24x28x96    &        \\
			Concat          &                & \begin{tabular}[c]{@{}c@{}}24x28x96 \&\\ 24x28x96\end{tabular} &                                                                                      & 24x28x192   &        \\
			(1x1) 2d conv   &                & 24x28x192                                                       & 192,96                                                                               & 24x28x96    & 8      \\
			Output          & 24x28x96       &                                                                 &                                                                                      &             &        \\ \bottomrule
		\end{tabular}
	}
	\label{tab:arch optical flow}
\end{table}


\section{Incorporation of LGA in different architectures}
\label{sec:incorp. of LGA}

Our LGA module takes a feature map $F^{\rm in}$ of size $H \times W \times C$ as input, where $H$, $W$, $C$ are height, width and number of channels respectively. It then returns an output feature $F^{\rm out}$ of the same size i.e. $H \times W \times C$ in which each feature of size $1 \times 1 \times C$ also contains the information of other spatially closed features. Later, $F^{\rm in}$ and $F^{\rm out}$ are concatenated as the final output of LGA, $F^{\rm cat}$. 

Example schematics of incorporating LGA into existing networks is shown in Fig. \ref{fig:teaser}. We pass an image $X$ to the model, the feature map generated by initial blocks of the model (or the network preceding the LGA) is passed as the input feature $F^{\rm in}$ to LGA and the concatenated output of LGA, i.e. $F^{\rm cat}$, is passed into the later blocks of the model (or the network succeeding LGA). 
For example, for a segmentation model consisting of an encoder-decoder architecture, the feature map generated by encoder can be considered as $F^{\rm in}$ and the $F^{\rm cat}$ is passed to the decoder block. Incorporation of LGA module is not limited to encoder-decoder kind of networks only. The LGA module can be inserted between any two blocks in an architecture as suitable. In this paper, we have presented three example applications, \textit{viz.} segmentation of transparent objects, image dehazing and optical flow estimation. In the following sections, we present the details of how we incorporated LGA into the architectures solving these three vision problems.

\subsection{LGA adaptation for semantic segmentation}
Our network for semantic segmentation is shown in Fig. \ref{fig:teaser}'s top panel. Our network consists of three blocks, namely an encoder, an LGA module and an up-sample block.
We removed some of the last layers of the encoder and did adequate padding (where needed) so that, given the input image of shape $H \times W \times 3$, the output is a feature map of size $H' \times W' \times C$. This feature map is then passed into LGA and the output is concatenated with the input of LGA. Then this concatenated feature map is passed to up-sample block. This block upscales the input feature map to generate a mask of shape $H \times W \times C_1$, where $C_1$ is the total number of classes. We have used group transposed convolution in the up-sample block for efficient up-sampling. For segmentation, we adapted LGA into SqueezeNet \cite{Squeeze1} and ShuffleNet \cite{zhang2017shufflenet}.

\subsection{LGA incorporation in BPPNet~\cite{singh2020single} for image restoration task}
The network for image restoration task is given in Fig. \ref{fig:teaser}'s middle panel. We adapted BPPNet ~\cite{singh2020single} by incorporating LGA with single UNet instead recurrent UNet by original BPPNet. We have incorporated our LGA module in between a UNet and pyramid convolutional network. We did not used 2d conv layer to change the input channel size because the size was already low (16). We added an additional 2D convolution layer of size $1 \times 1$ after LGA to make the channel size of the concatenated output feature equal to the channel size of LGA's input feature map. We included this layer because the pyramid block does maximum number of operations and it is proportional to the input size. Hence, maintaining the same channel size gives better computational efficiency. All the convolutional layers of UNet and pyramid convolutional block were replaced by group convolutions in order to further improve the overall computational efficiency.

\subsection{LGA incorporation in ARFlow \cite{liu2020learning} for optical flow estimation}
The network for optical flow estimation task is given in Fig. \ref{fig:teaser}'s lower panel. We used ARFlow \cite{liu2020learning} as our base model for unsupervised optical flow estimation. ARFlow extracts the feature pyramid of the frames and obtains a flow map via the correlation between feature maps and uses the convolution layer for refining it. We believe that if we can extract much better features, which have information of its neighbors, we will be able to improve the flow map. For this reason, we added our LGA module after output feature extractor module. In our experiment, we only applied LGA at a single feature map of spatial size (24, 28). This is because the smaller maps don’t have much spatial information because of their smaller size, which becomes a challenge for LGA to propagate information and enhance the map. For higher maps, because the information of smaller scale features are used for high scale flow generation via using small scale flow map as prior, applying LGA at latter stages does not make much difference and it also increases the overall time complexity. When applying LGA feature of size (24,28) we did not apply 2d Conv layer for channel resizing because the input size was optimal $(96)$. We also, applied a 2d Conv layer after concatenating the input and output map, so as to make the input channel size and final output channel size equal.

\bibliographystyle{unsrt}  






\end{document}